\documentclass[10pt,twocolumn,letterpaper]{article}

\usepackage[pagenumbers]{cvpr} 
\usepackage{multirow}
\usepackage{xcolor}
\usepackage{colortbl}
\usepackage[export]{adjustbox}
\usepackage{float}
\definecolor{cvprblue}{rgb}{0.21,0.49,0.74}
\usepackage[pagebackref,breaklinks,colorlinks,allcolors=cvprblue]{hyperref}
\newcommand{\bfsection}[1]{\vspace*{0.1cm}\noindent\textbf{#1.}}
\def\paperID{14927} 
\def\confName{CVPR}
\def\confYear{2026}

\title{SparseWorld-TC: Trajectory-Conditioned Sparse Occupancy World Model}

\author{
  Jiayuan Du$^{1,2}$\thanks{Equal contribution},
  Yiming Zhao$^{2}$\footnotemark[1],
  Zhenglong Guo$^{2}$,
  Yong Pan$^{2}$,
  Wenbo Hou$^{2}$,
  Zhihui Hao$^{2}$, \\
  Kun Zhan$^{2}$,
  Qijun Chen$^{1}$\thanks{Corresponding author} \\
  $^1$Tongji University, $^2$Li Auto Inc. \\
  {\tt\small dujiayuan@tongji.edu.cn, qjchen@tongji.edu.cn}
}

\begin{document}
\maketitle
\begin{abstract}

This paper introduces a novel architecture for trajectory-conditioned forecasting of future 3D scene occupancy. In contrast to methods that rely on variational autoencoders (VAEs) to generate discrete occupancy tokens, which inherently limit representational capacity, our approach predicts multi-frame future occupancy in an end-to-end manner directly from raw image features. Inspired by the success of attention-based transformer architectures in foundational vision and language models such as GPT and VGGT, we employ a sparse occupancy representation that bypasses the intermediate bird's eye view (BEV) projection and its explicit geometric priors. This design allows the transformer to capture spatiotemporal dependencies more effectively. By avoiding both the finite-capacity constraint of discrete tokenization and the structural limitations of BEV representations, our method achieves state-of-the-art performance on the nuScenes benchmark for 1‒3 second occupancy forecasting, outperforming existing approaches by a significant margin. Furthermore, it demonstrates robust scene dynamics understanding, consistently delivering high accuracy under arbitrary future trajectory conditioning. Code: \href{https://github.com/MrPicklesGG/SparseWorld-TC}{https://github.com/MrPicklesGG/SparseWorld}.

\end{abstract}    
\section{Introduction}
\label{sec:intro}

World models offer a principled framework for understanding environmental dynamics and are considered essential for AI systems operating in the physical world. Although the literature has yet to converge on a unified and explicit definition of world models, the ability to predict the evolution of the physical environment is consistently regarded as a core capability \cite{suervey_1, suervey_2, li2025scaling}. Among recent advances, occupancy-based world models have emerged as a prominent category \cite{occsora, occworld, occ3d, occ-llm, i2world, dome, come}, garnering significant interest due to their direct applicability in autonomous driving and robotics, as well as their role in generating temporally consistent sensor observations \cite{dynamiccity, infinicube}.

\begin{figure}[t]
  \centering
   \includegraphics[width=1.0\linewidth]{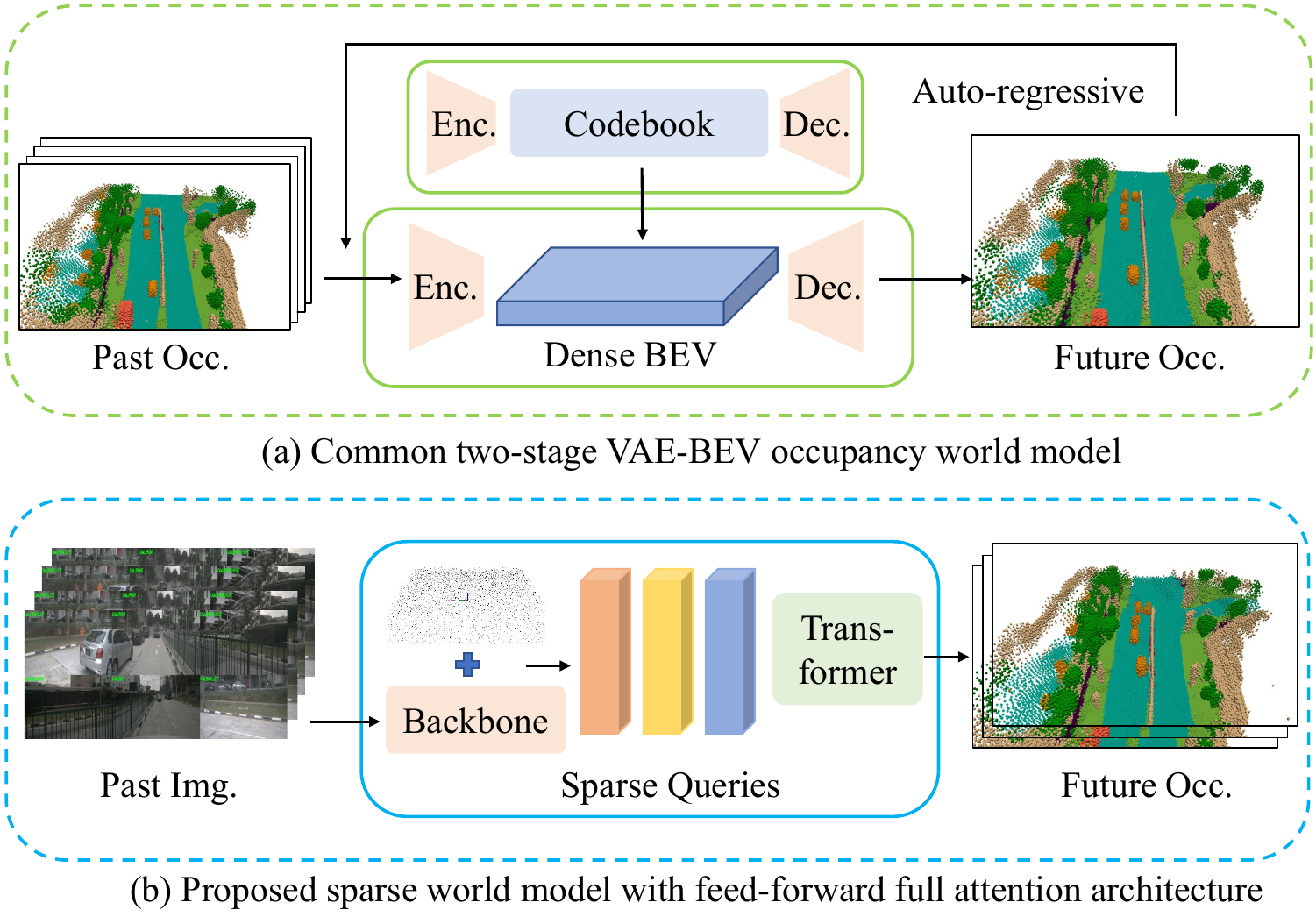}
   \caption{Without VAE codebook and BEV representations, the model represents future multi-frame occupancy using sparse queries and leverages image evidence from prior frames to convert randomly sampled 3D points into reliable future occupancy.}
   \label{fig:intro}
\end{figure}

Despite this progress, many recent approaches encode observations into discrete tokens using variational autoencoders (VAEs) \cite{vae, vqvae} and then predict future occupancy from these limited-vocabulary representations \cite{occworld, come, dome, i2world}. Such tokenization can constrain representational capacity and discard fine-grained information. Alternatively, some methods recast occupancy world modeling as predicting dynamic and static flow fields \cite{uniocc, dfit-occworld}, which does not inherently preserve generative capabilities.

\bfsection{Motivation} Pure attention-based feed-forward architectures have demonstrated remarkable flexibility and representational capacity in both language and 3D vision domains, as exemplified by models such as GPT \cite{gpt4} and VGGT \cite{vggt}. However, these capabilities remain underexplored for future occupancy prediction. Unlike methods that depend on fixed-resolution bird's eye view (BEV) representations, which introduce explicit geometric constraints into feature interactions, we propose a pure attention-based transformer architecture that leverages fully sparse occupancy representations to directly capture spatiotemporal relationships from raw image features. Our approach eliminates both discretized tokens and explicit BEV geometric priors, enabling more effective integration of spatiotemporal dependencies, as illustrated in Figure~\ref{fig:intro}.

Building upon recent advances in sparse perception models \cite{sparsebev, sparsedrive}, we represent scene occupancy using a set of anchors, each composed of a group of 3D query points and an associated feature vector. For each future frame, the anchor features interact through self-attention and cross-attention mechanisms with image features from past frames, accessed via deformable attention. This process iteratively refines the initially randomized points into accurate, context-aware occupancy representations. Trajectory information is incorporated as a conditioning signal to steer the temporal evolution of the predicted scene.

\bfsection{Contribution} We summarize our contributions below:
\begin{itemize}
\item We propose a pure attention-based transformer architecture for occupancy world modeling, which integrates spatiotemporal information through sparse occupancy representations in an end-to-end manner, eliminating the need for discrete tokens and bypassing BEV geometric priors.
\item We introduce a trajectory-conditioned occupancy forecasting framework that facilitates precise guidance of scene evolution along given trajectories while ensuring high temporal consistency.
\item Our method achieves state-of-the-art performance on the nuScenes benchmark \cite{nuscenes, occ3d} for 1--3 second occupancy forecasting and demonstrates compelling results on long-term forecasting benchmarks \cite{come, dome}. 
\end{itemize}
\section{Related Work}
\label{sec:related_work}


4D occupancy forecasting, which models spatiotemporal scene occupancy (3D space + 1D time) from historical observations, serves as a core foundation for autonomous driving. Recent advances in this area can be categorized along three key technical dimensions: tokenization strategy, intermediate representation , and generation paradigm.

For \textbf{tokenization strategies}, existing works \cite{occworld, occsora, fsfnet, occllama, dome, renderworld, occ-llm, occvar, i2world, come} mainly adopt VAE \cite{vae} or VQ-VAE \cite{vqvae} based approaches. OccWorld \cite{occworld} utilizes an occupancy VQ-VAE to discretize continuous 3D scene data into discrete tokens. OccLLaMA \cite{occllama} extends this direction with a VAE-like scene tokenizer optimized for the sparsity and class imbalance of semantic occupancy, enabling unified visual-language-action modeling. RenderWorld \cite{renderworld} introduces AM-VAE, a dual-codebook variant of VQ-VAE that separately encodes air and non-air voxels to enhance the granularity of the representation. I$^2$-World \cite{i2world} further advances VAE-based tokenization by designing a multi-scale architecture, which refines token granularity across different spatial scales to capture both local details and global scene structures. In contrast, DOME \cite{dome} and COME \cite{come} employ a continuous Occ-VAE instead of discrete tokenization, preserving fine-grained details through continuous latent spaces.

\textbf{Intermediate representations} must balance precision and efficiency, including BEV feature \cite{occworld, dfit-occworld, occvar, preworld, i2world, dome, come, occprophet, efficientocf}, Gaussian representation \cite{gaussianad, renderworld}, and others like triplane \cite{dtt} or hexplane \cite{dynamiccity}. Most VAE-based methods \cite{occworld, dome, come, occvar, i2world, occllama} rely on dense BEV feature maps for spatial-temporal modeling and conditional scene generation. PreWorld \cite{preworld} validates BEV's effectiveness through pre-training, while DFIT-OccWorld \cite{dfit-occworld} uses it for flow prediction. GaussianAD \cite{gaussianad} and RenderWorld \cite{renderworld} adopt 3D Gaussian splatting \cite{3dgaussian} to model scenes as anisotropic Gaussian ellipsoids, achieving efficient rendering and high segmentation accuracy. DTT \cite{dtt} and DynamicCity \cite{dynamiccity} utilize triplane or hexplane as intermediate representations, which retain 3D structural information while achieving more compact latent spaces.

Regarding \textbf{generation paradigms}, autoregressive modeling \cite{occworld, preworld, renderworld, occvar, i2world, drive-occworld, occ-llm, occllama} remains prevalent for sequential scene prediction, though non-autoregressive \cite{dfit-occworld} and diffusion-based \cite{dtt, dome, come} alternatives have gained traction. OccWorld \cite{occworld} and RenderWorld \cite{renderworld} use hierarchical transformers for autoregressive future scene token prediction. OccLLaMA \cite{occllama} and Occ-LLM \cite{occ-llm} leverage large language models to perform cross-modal next-token/scene autoregressive modeling. DFIT-OccWorld \cite{dfit-occworld} stands out by adopting a non-autoregressive paradigm based on its BEV intermediate representations, which accelerates generation speed compared to autoregressive counterparts while maintaining scene integrity. DOME \cite{dome} leverages a spatial-temporal diffusion transformer, and COME \cite{come} extends this diffusion framework with ControlNet \cite{controlnet} to enable conditional diffusion-based generation.

\bfsection{Proposed Method} Rather than relying on handcrafted tokenizers or intermediate representations, our approach prefers to model the occupancy-based world in an end-to-end manner. Scenes are represented as sets of learnable feature embeddings, with their interactions mediated by attention mechanisms. This formulation departs from diffusion-based and autoregressive paradigms, favoring a VGGT-like feed-forward architecture \cite{vggt} that predicts future occupancy in a single forward pass. Under this design, conditioning scene evolution on a given trajectory is straightforward: we embed the trajectory as feature vectors and let them interact with the scene features.

\section{Methodology}
\label{sec:method}
\subsection{Sparse World Model Representation}
\begin{figure}[t]
  \centering
   \includegraphics[width=1.0\linewidth]{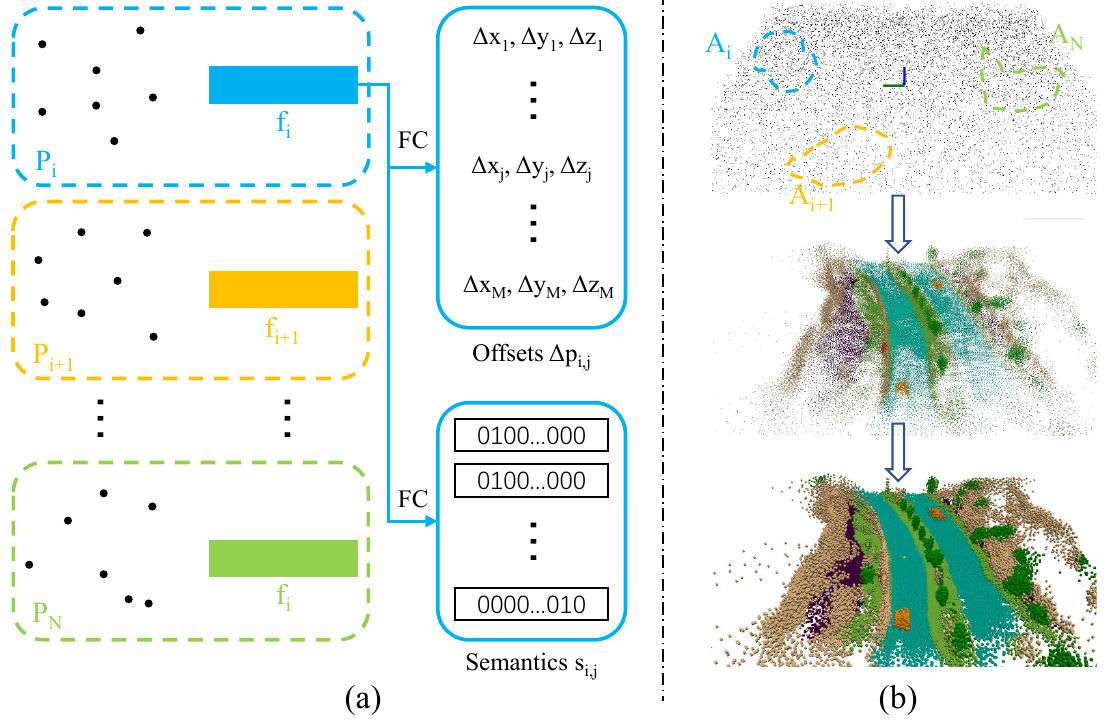}

   \caption{Occupancy is modeled as a collection of anchors formed by a bundle of 3D points paired with an occupancy embedding. The embedding is decoded by two MLPs to yield per point offsets and semantic class logits, which denoise sampled points to a consistent occupancy field.}
   \label{fig:representation}
   \vspace{-1.0em}
\end{figure}

Most existing approaches model 3D scene occupancy using dense voxel grids~\cite{fbocc}, which are typically constructed on BEV representations. Although effective, such BEV-based paradigms impose explicit geometric constraints that can limit the flexible interaction of spatiotemporal features across different scales. Inspired by recent advances in sparse perception models that bypass BEV as an intermediate representation~\cite{petrv2, sparse4d, sparsebev, opus}, we propose an alternative occupancy formulation based on a collection of anchors. In our framework, each anchor consists of a set of randomly initialized 3D points and an associated feature vector that jointly predicts per-point geometric offsets and semantic labels, as illustrated in Figure~\ref{fig:representation}. Based on this sparse occupancy representation, we define the world model as a sequence of such single-frame occupancy states conditioned on a given trajectory.

\bfsection{Single Frame Occupancy Representation} 
We represent the initial occupancy of a single frame by $\mathcal{M}$, which comprises $N$ anchors, each associated with a feature vector:
\begin{equation} \label{eq1}
\mathcal{M} = \{ A_i \}_{i=1}^{N}, \quad A_i = (P_i, \mathbf{f}_i)
\end{equation}
where $P_i = \{ \mathbf{p}_{i,j} \in \mathbb{R}^3 \}_{j=1}^{M}$ denotes a set of $M$ 3D points for anchor $i$, and $\mathbf{f}_i \in \mathbb{R}^D$ is the corresponding feature vector. The feature vector $\mathbf{f}_i$ predicts for each point $\mathbf{p}_{i,j}$ both an offset $\Delta \mathbf{p}_{i,j} \in \mathbb{R}^3$ and a semantic label $\mathbf{s}_{i,j} \in \mathbb{R}^C$ (a $C$-dimensional class probability vector). The representation $\mathcal{M}$ is initialized as follows: the centers $\{ \mathbf{c}_i \}_{i=1}^{N}$ are uniformly distributed in the 3D space, $\mathbf{c}_i \sim \mathcal{U}(\text{Space})$; each set of points $P_i$ is randomly initialized within a local region around $\mathbf{c}_i$ as $\mathbf{p}_{i,j} \gets \mathbf{c}_i + \boldsymbol{\epsilon}_{i,j}$ with $\boldsymbol{\epsilon}_{i,j} \sim \mathcal{N}(\mathbf{0}, \sigma^2\mathbf{I})$; and all feature vectors are initialized to zero, $\mathbf{f}_i \gets \mathbf{0}$.

\begin{figure*}[t]
  \centering
   \includegraphics[width=1.0\linewidth]{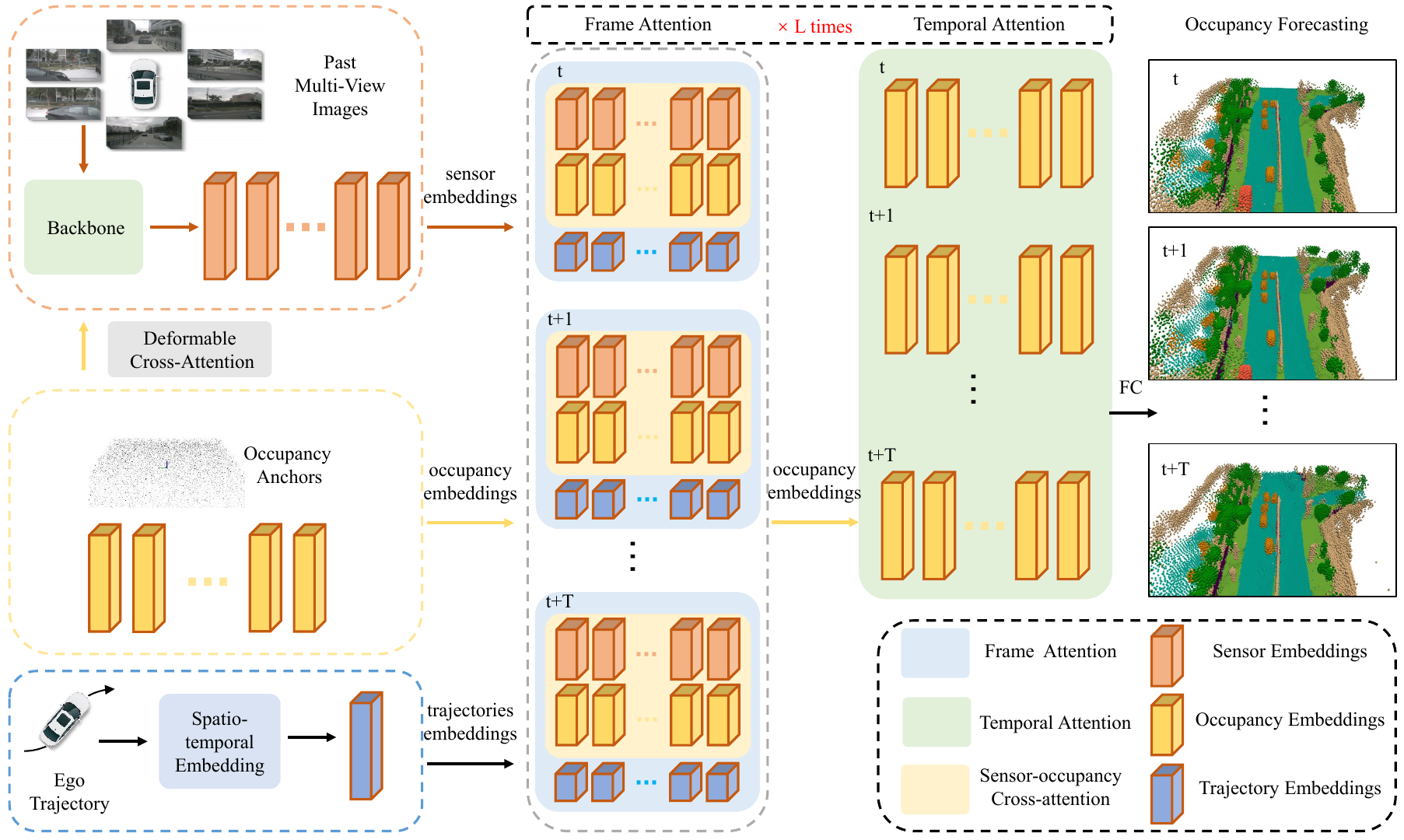}
   \caption{We embed the sensor observations, occupancy priors, and trajectories into feature vectors. These embeddings pass through stacked frame-level attention and temporal attention blocks that iteratively fuse cross-modal and cross-time information into occupancy features. Those occupancy features finally decode random points into meaningful occupancy predictions.}
   \label{fig:overview}
   \vspace{-1.0em}
\end{figure*}

\bfsection{Trajectory Representation} 
In autonomous driving, the planned future trajectory of the ego vehicle provides an essential conditioning signal for predictive world modeling. We parameterize a future trajectory $\tau$ as a sequence of discrete states spanning a temporal horizon $T$:

\begin{equation} \label{eq2}
\tau = \{ \mathbf{x}_t \}_{t=1}^{T}
\end{equation}

Each state $\mathbf{x}_t \in \mathbb{R}^d$ encapsulates the ego's kinematic status at time $t$. In our formulation, each state comprises the vehicle's planar position $(x, y)$, its heading angle $\theta$, and the timestamp $t$ itself, providing a compact yet expressive representation for conditioning the world model.

\bfsection{World Model Representation} 
The world model $\mathcal{F}$ synthesizes future scene evolution conditioned on the planned trajectory and historical observation. Formally, the model generates a sequence of future occupancy states:

\begin{equation} \label{eq3}
\mathcal{O}_{1:T} = \mathcal{F}\left( \mathcal{M}_{1:T}, \mathbf{S}_{-T':0}, \tau \right)
\end{equation}

where $\mathcal{M}_{1:T}$ denotes the initial state representation of all future frames, $\mathbf{S}_{-T':0}$ represents a history of sensor observations from the past $T'$ timesteps, and $\tau$ is the given trajectory. This formulation enables the model to integrate past context and future intentions to render physically consistent future scenes, thereby supporting downstream tasks such as motion planning and predictive safety assessment. Since the anchor points in $\mathcal{M}_{1:T}$ are stochastically initialized, our occupancy world model operates similarly to a conditional video generation model that transforms random 3D point clouds into sequential driving scenarios conditioned on both past sensor observations and future trajectory information.

\begin{figure*}[ht]
\centering
\footnotesize
\setlength{\tabcolsep}{0.05cm}
\newcolumntype{P}[1]{>{\centering\arraybackslash}m{#1}}
\scalebox{0.99}{
\begin{tabular}{P{0.7cm}P{4cm}P{4cm}P{4cm}P{4cm}}
 & Recon. & 1s & 2s & 3s \\

\rotatebox[origin=c]{90}{(a) GT} & {\includegraphics[width=\linewidth, frame]{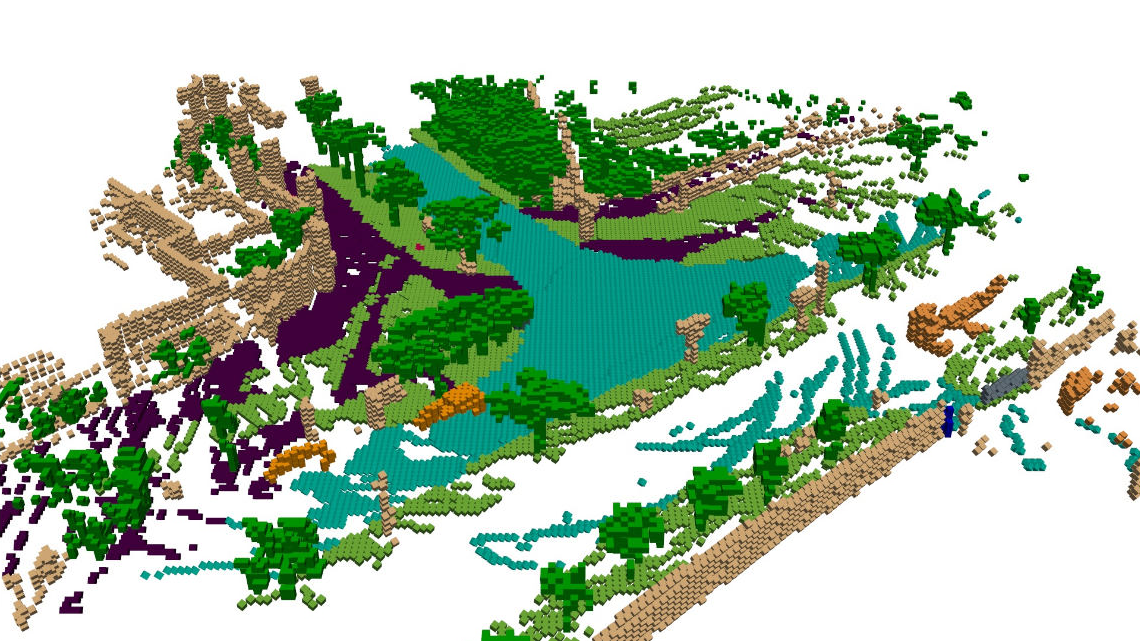}} & {\includegraphics[width=\linewidth, frame]{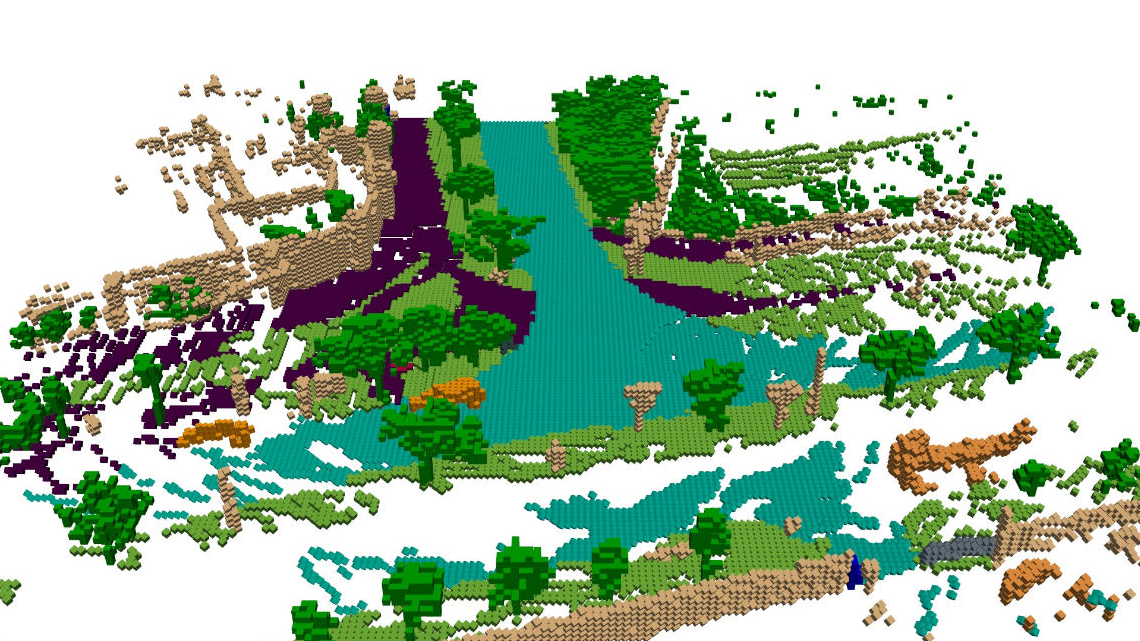}} & {\includegraphics[width=\linewidth, frame]{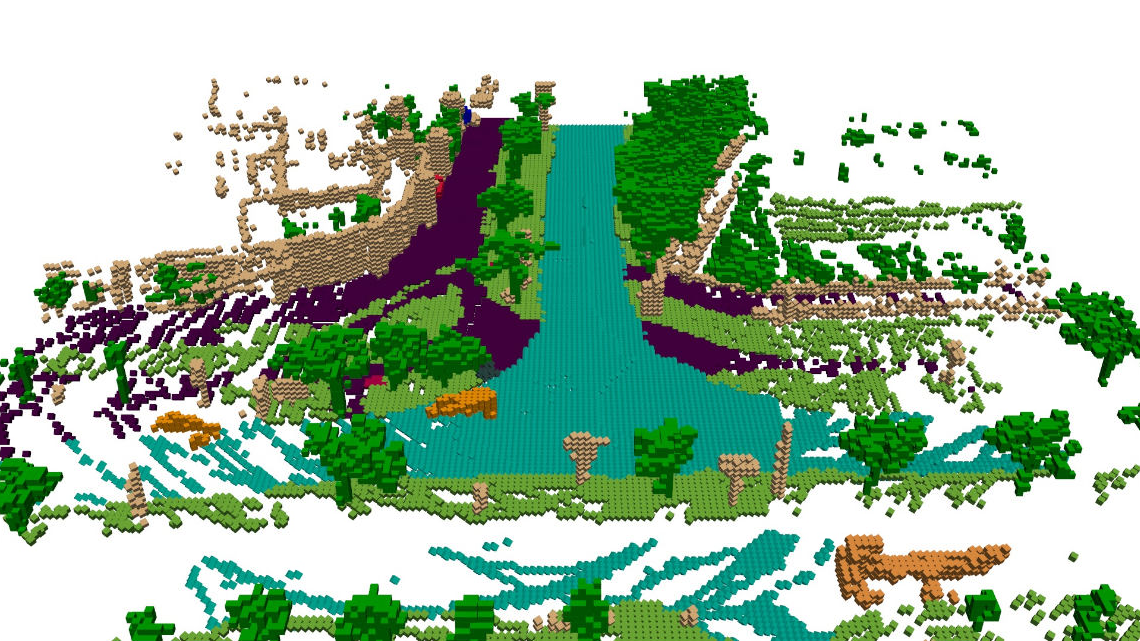}} & {\includegraphics[width=\linewidth, frame]{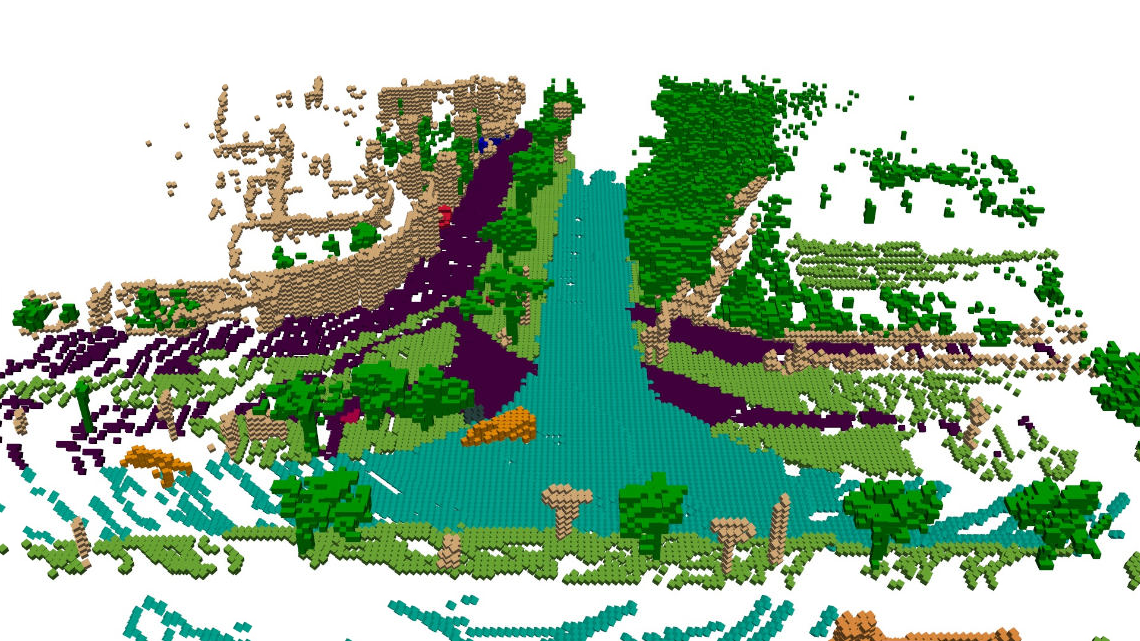}} \\

\rotatebox[origin=c]{90}{(a) Pred.} & {\includegraphics[width=\linewidth, frame]{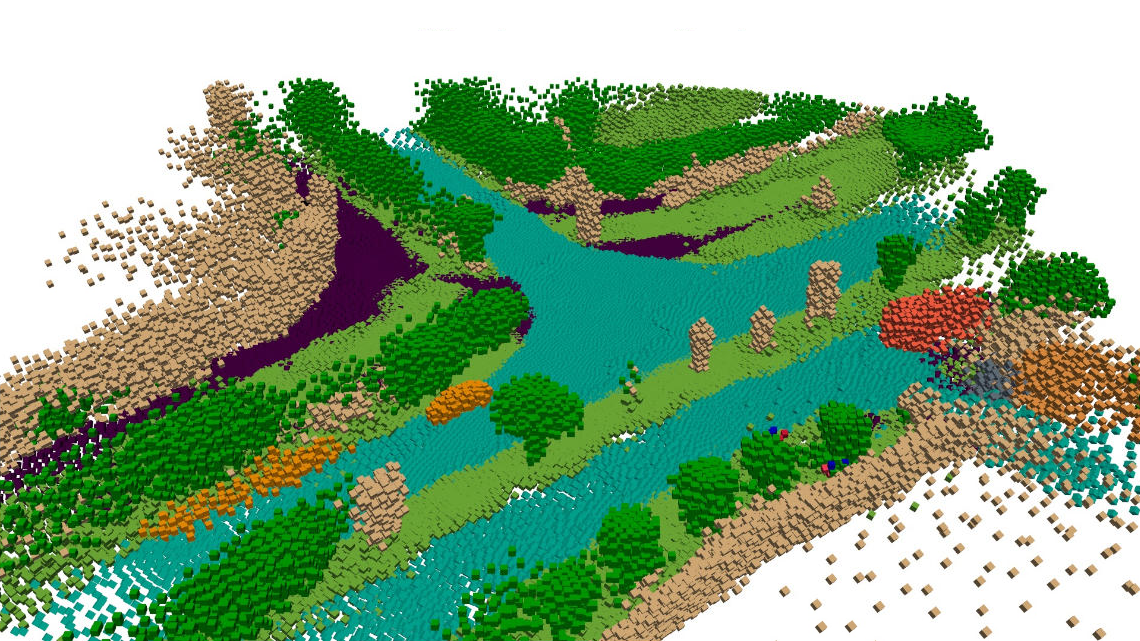}} & {\includegraphics[width=\linewidth, frame]{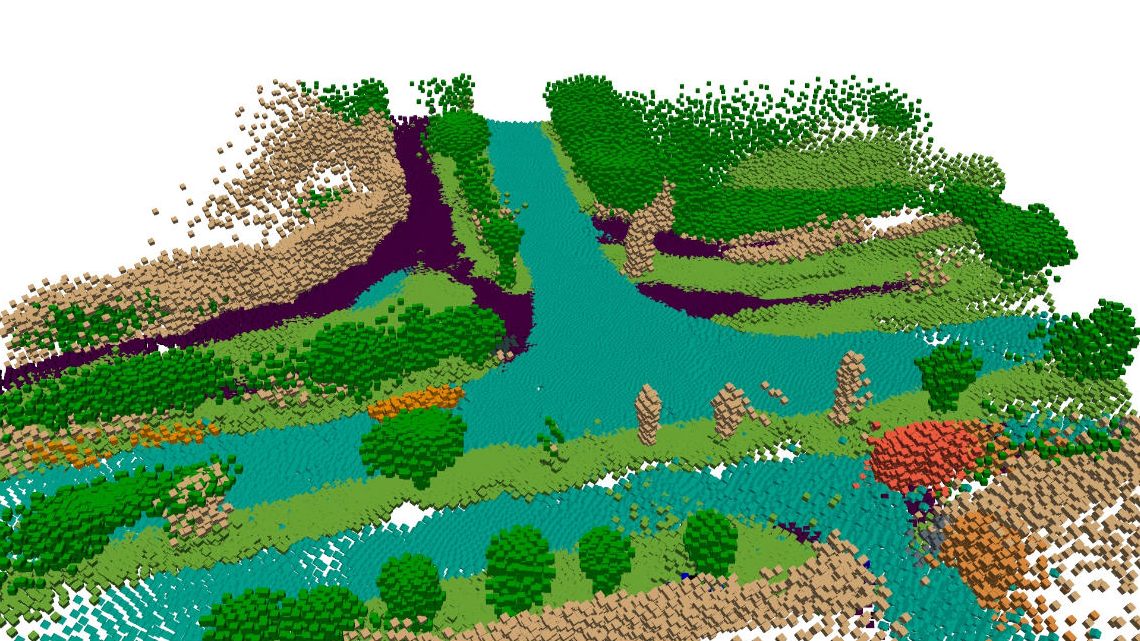}} & {\includegraphics[width=\linewidth, frame]{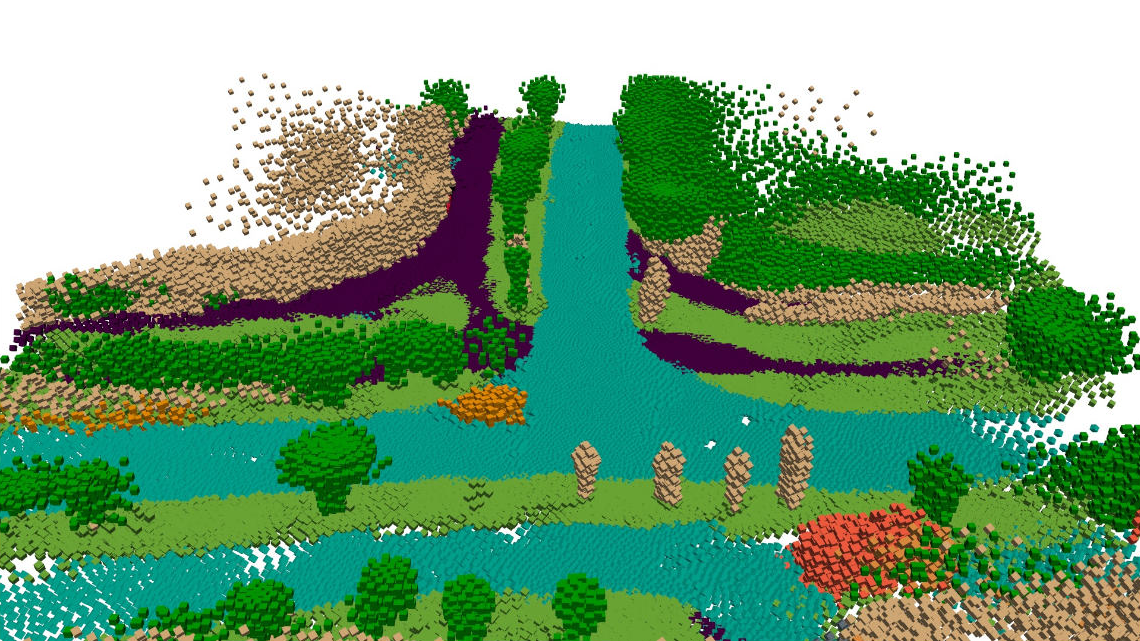}} & {\includegraphics[width=\linewidth, frame]{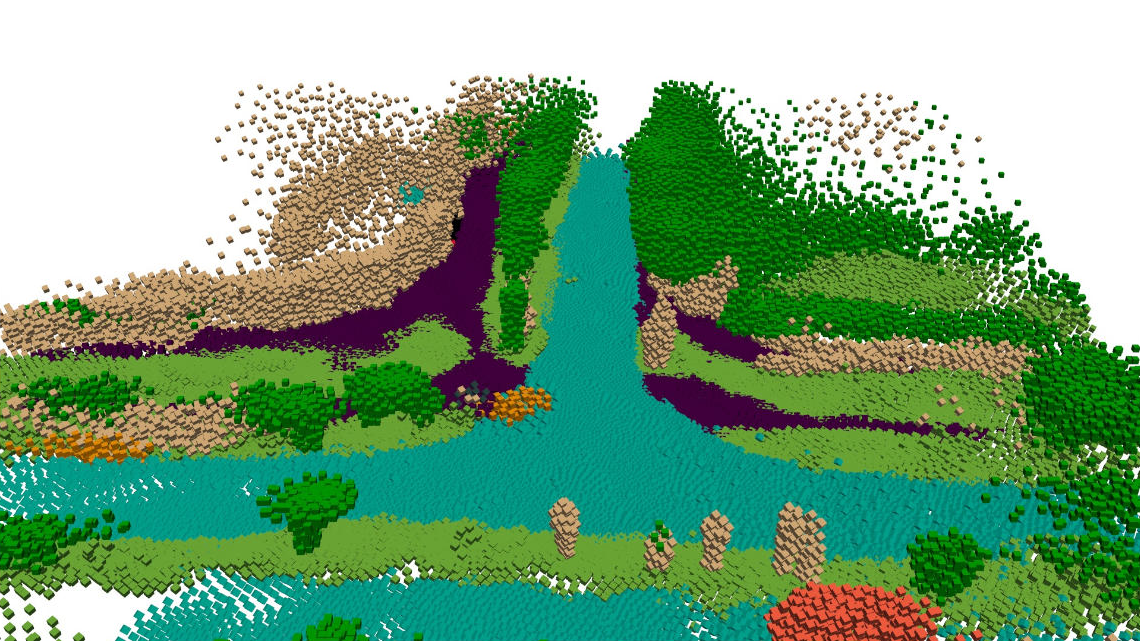}} \\

\rotatebox[origin=c]{90}{(b) GT} & {\includegraphics[width=\linewidth, frame]{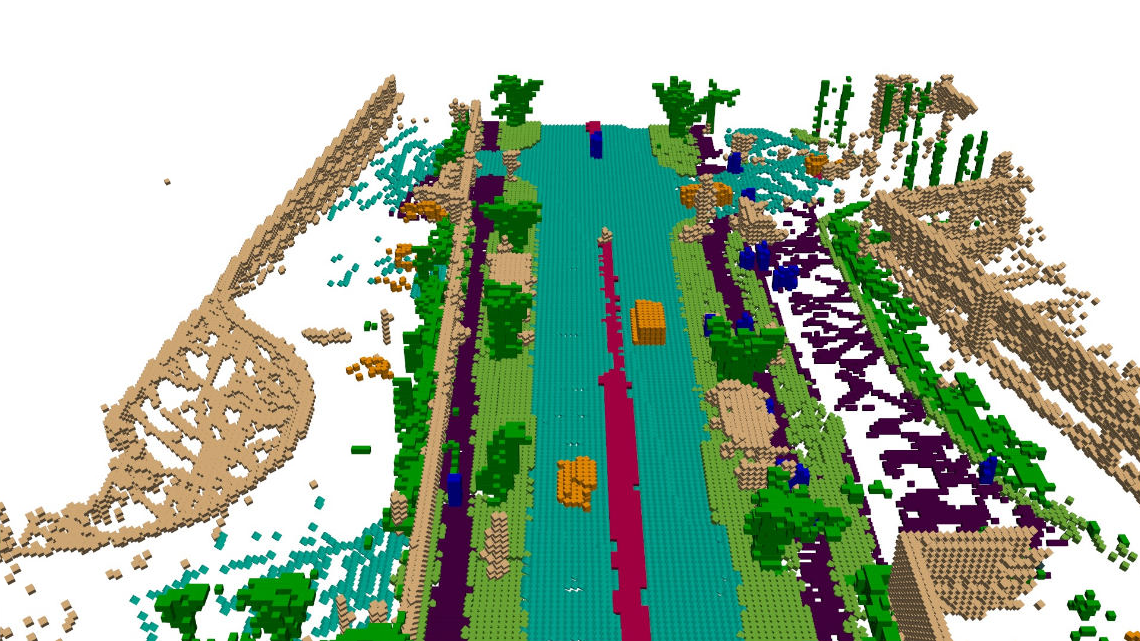}} & {\includegraphics[width=\linewidth, frame]{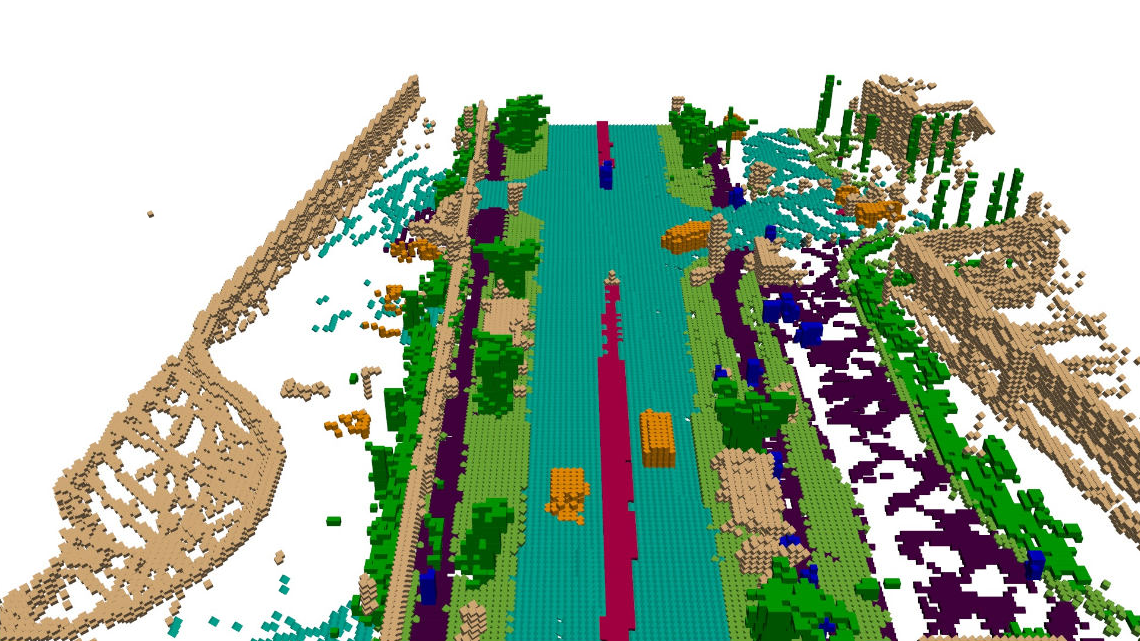}} & {\includegraphics[width=\linewidth, frame]{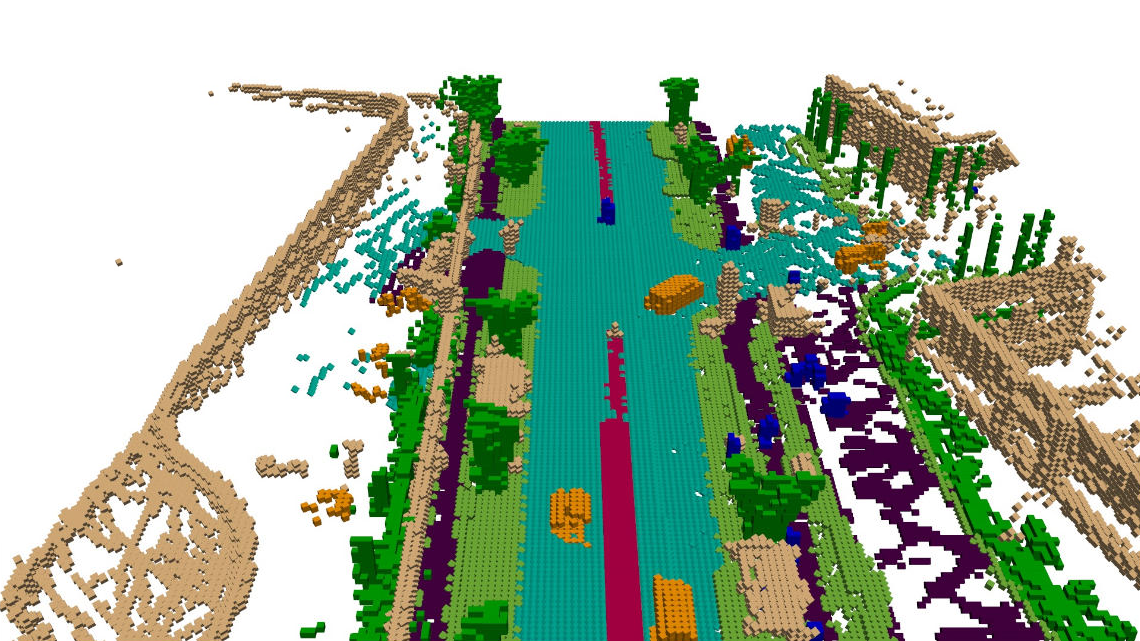}} & {\includegraphics[width=\linewidth, frame]{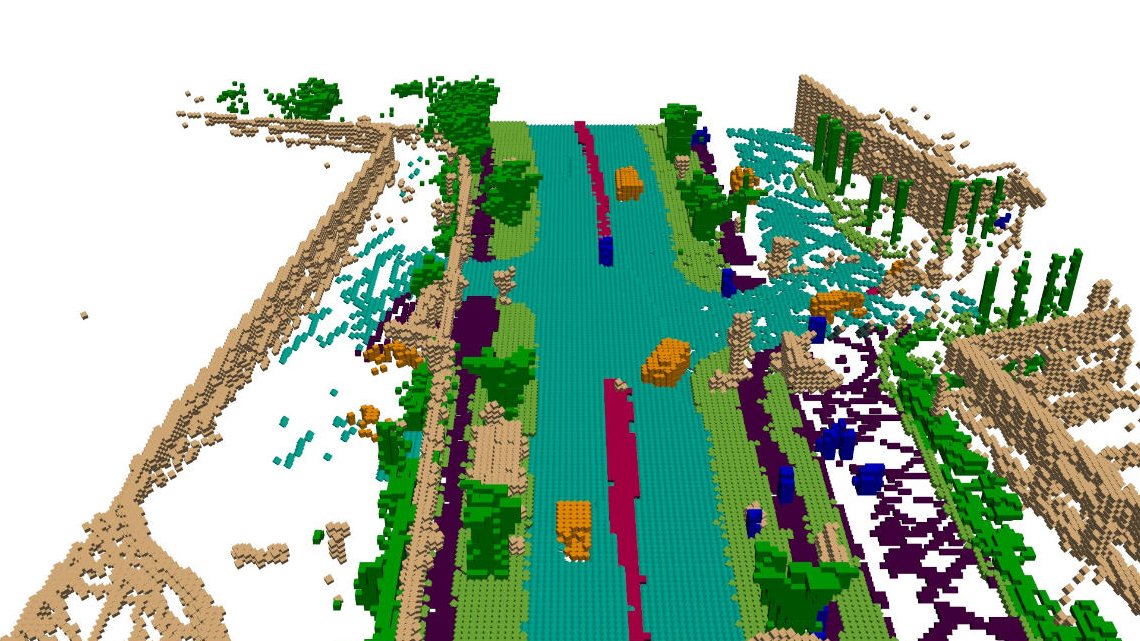}} \\

\rotatebox[origin=c]{90}{(b) Pred.} & {\includegraphics[width=\linewidth, frame]{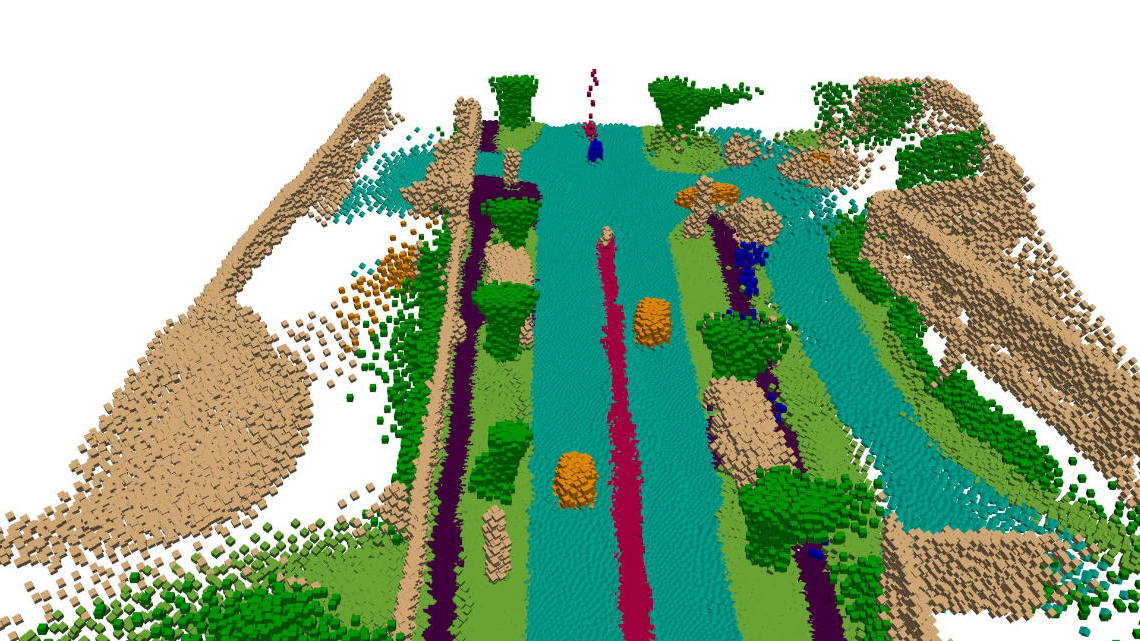}} & {\includegraphics[width=\linewidth, frame]{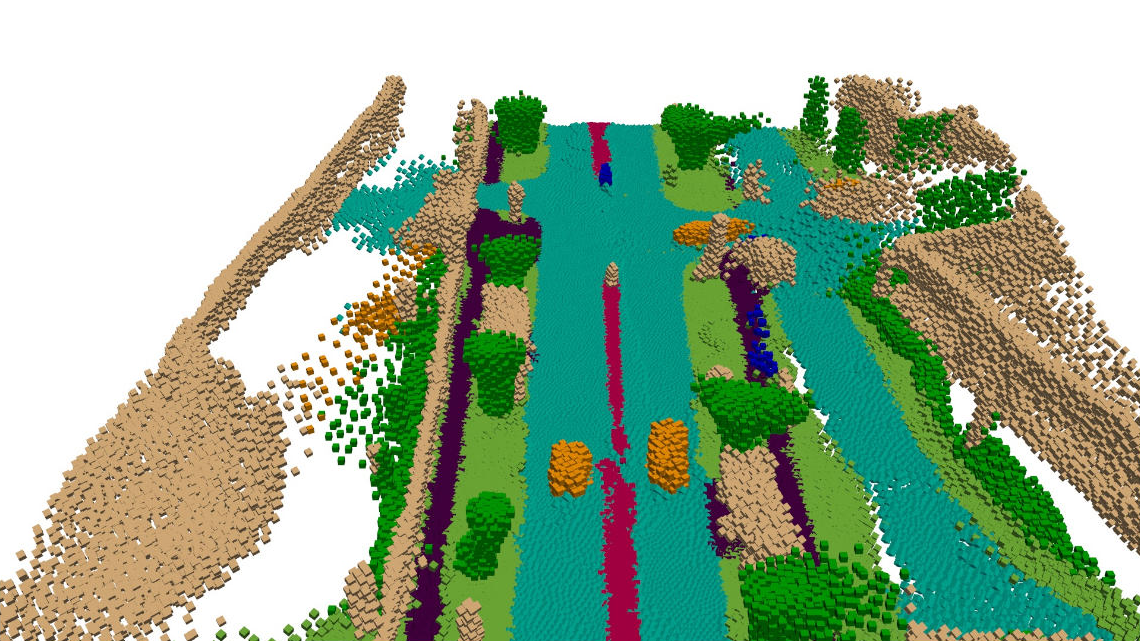}} & {\includegraphics[width=\linewidth, frame]{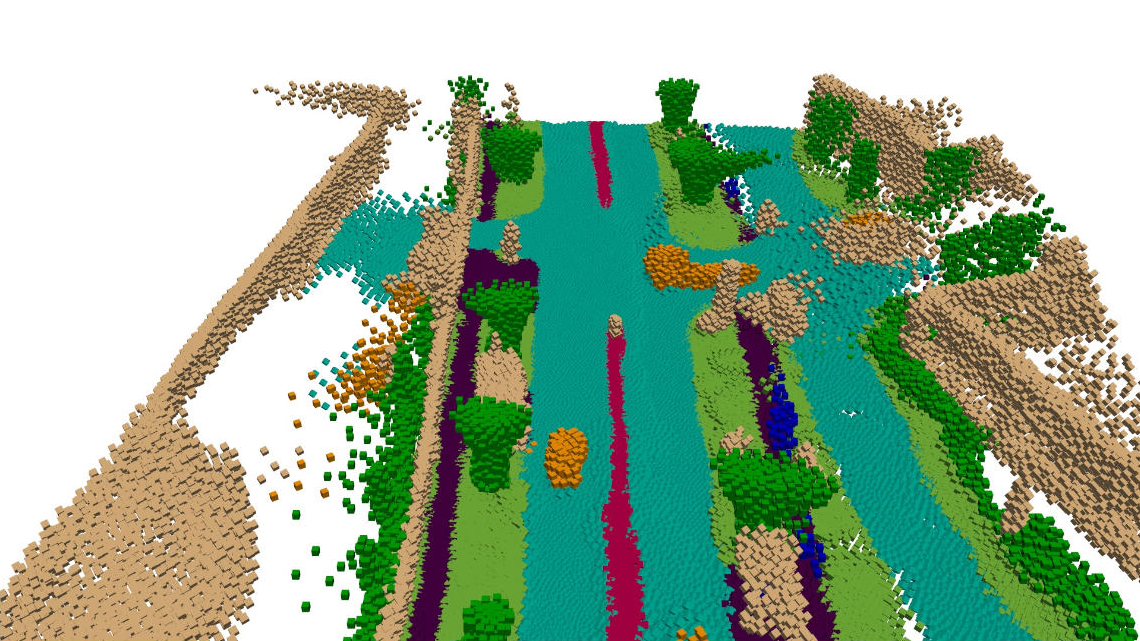}} & {\includegraphics[width=\linewidth, frame]{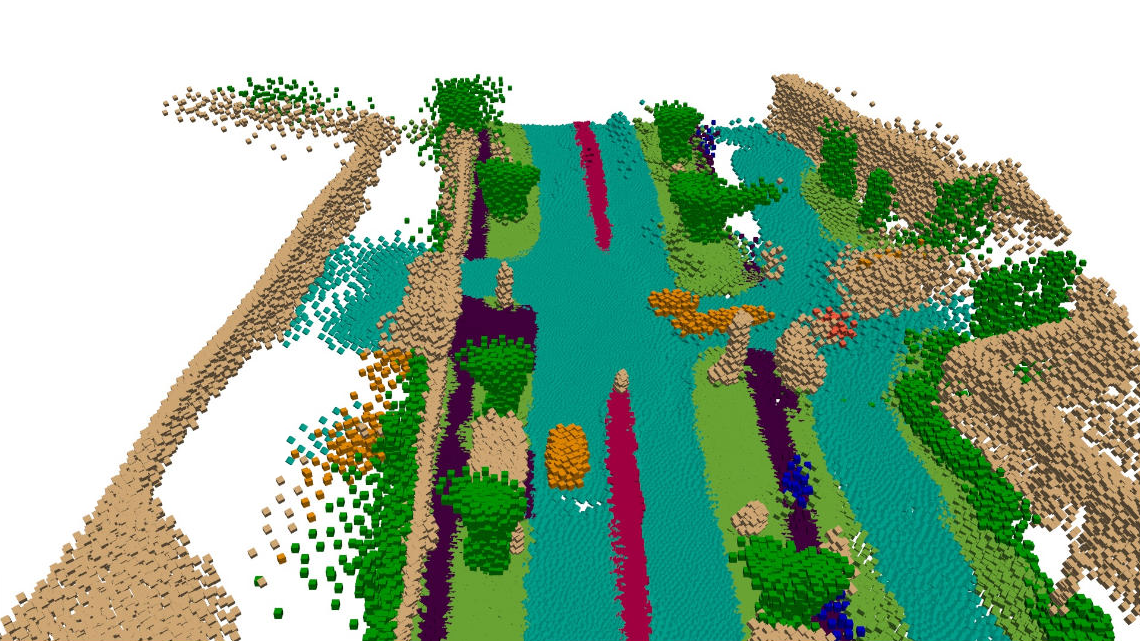}} \\

\end{tabular}
}
\caption{Qualitative results of our proposed SparseWorld-TC are presented here. Our method effectively captures both dynamic and static changes in the surrounding environment over a 3-second forecasting horizon.}
\vspace{-1.0em}
\label{fig:vis_nusc}
\end{figure*}

\subsection{Spatial Temporal Fusion Architecture}
In Equation~\ref{eq1}, the occupancy of the scene is represented by a set of feature vectors. We further embed both the trajectory waypoints and the historical sensor observations into separate feature representations. Once all modalities are projected into a unified embedding space, they can be efficiently fused and interact through attention mechanisms within our proposed pure attention-based architecture.

\subsubsection{Trajectory Spatiotemporal Embedding}

As mentioned above, our trajectory representation incorporates the position and timestamp of each waypoint. To maintain flexibility, we do not assume uniform temporal sampling between waypoints. Instead, we directly embed the individual position and timestamp of each waypoint to accommodate various potential planning outputs.

\bfsection{Position Embedding} Each waypoint is defined by a 7-dimensional global coordinate \((x, y, z, q_0, q_1, q_2, q_3)\) and transformed to a 16-dimensional homogeneous matrix (relative coordinate in the current ego). The positional feature vector \(\mathbf{f}_i^{P_n}\) is obtained by an MLP that projects the 16-dimensional matrix to the same dimension as the corresponding feature vector \(\mathbf{f}_i\) for occupancy anchor \(i\).

\bfsection{Time Embedding} Under the assumption of a fixed raw sensor frequency, the timestamps can be represented as \([0, 1, \dots, n]\), where \(0\) denotes the current timestamp. For timestamp \(n\), this integer is encoded into a positional embedding \(\mathbf{f}_i^{T_n}\) using standard sinusoidal positional encoding.

\bfsection{Spatiotemporal Embedding} Drawing inspiration from motion-aware layer normalization (MLN) \cite{streampetr}, spatiotemporal embedding achieves the fusion of spatial and temporal information by building on the aforementioned position embedding and time embedding. Specifically, two linear layers are designed to implicitly learn the affine transformations between adjacent frames. The spatiotemporal embedding method is formulated as: $\mathbf{f}''_t = \gamma (\mathbf{f}_i^{P_n}) \cdot (\mathbf{f}_i + \mathbf{f}_i^{T_n}) + \beta (\mathbf{f}_i^{P_n})$, where both \(\gamma\) and \(\beta\) are distinct linear layers applied to \(\mathbf{f}_i^{P_n}\). The final trajectory $\tau$ is thus represented as a set of spatiotemporal feature vectors $ \tau = \{ \mathbf{f}''_t \}_{t=1}^T$.

\subsubsection{Sensor Embedding with Deformable Attention}
Deformable attention is widely used in 3D sparse perception \cite{detr3d, bevformer, sparse4d, sparsebev, sparsedrive}. It projects 3D query points into multi-view image feature maps (and 3D LiDAR voxel grids when LiDAR is available) using known calibration and ego-motion, and samples features at the projected locations.

In our architecture, we take into account $m$ sensors per frame, the center $\mathbf{c}_i$ of each anchor point set $P_i$ serves as the query point in the deformable attention operation, where we calculate the mean and standard deviation of anchor point set $P_i$ along the $x, y, z$ directions as the basis for sampling offsets. Each center is projected into the multi-scale image feature maps of the backbone (e.g. ResNet \cite{resnet} or ViT \cite{vit}) using camera intrinsics, extrinsics and the ego pose. If a query projects into multiple views due to field-of-view overlap, we aggregate by averaging the sampled features across all $m$ views. Each anchor center collects features from the past $T'$ frames. To encode temporal context, we add a sinusoidal time embedding processed by a fully connected layer to provide motion cues. Finally, for each future frame with occupancy representation $\mathcal{M}$, we obtain the sensor embedding representation $\mathcal{S} = \{\mathbf{f}'_i\}_{i=1}^{N \times T'}$.

\subsubsection{Fully Attention Fusion Architecture}

In Equation~\ref{eq3}, $\mathcal{F}\left( \cdot \right)$ denotes the neural network architecture to be designed. In our approach, the world occupancy representation $\mathcal{M}_{1:T}$, the sensor representation $\mathbf{S}_{-T':0}$, and the trajectory representation $\tau$ are all projected into a unified embedding space. Accordingly, the world model can be reformulated as:

\begin{equation} \label{eq4}
\mathcal{O}_{1:T} = \mathcal{F}\left( \{ \mathbf{f}_i \}_{i=1}^{N \times T}, \{ \mathbf{f}'_i \}_{i=1}^{N \times T \times T'}, \{ \mathbf{f}''_t \}_{t=1}^T \right)
\end{equation}
Note that the point sets $P_i$ in Equation~\ref{eq1} consist of non-learnable random coordinates and are therefore omitted from the input in Equation~\ref{eq4}.

Equation~\ref{eq4} provides a compact formulation of our world model: all relevant features interact directly via standard attention. We employ a feed-forward, purely attention-based transformer architecture, shown in Figure~\ref{fig:overview}. For each future frame 
$t$, the occupancy embeddings $\{ \mathbf{f}^{t}_{i} \}_{i=1}^{N}$ attend to the past sensor embeddings $\{ \mathbf{f'}_{i}^{t} \}_{i=1}^{N \times T'}$ via cross-attention. The updated occupancy features are then fused with the trajectory embedding through frame-level self-attention.  Finally, a temporal attention block applies self-attention across all future frames, jointly refining the set of occupancy embeddings $ \{ \mathbf{f}_{i} \}_{i=1}^{N \times T}$ to capture longer-range spatiotemporal dependencies. We stack the frame and temporal attention modules and apply them for multiple iterations, progressively refining the randomly initialized 3D anchor points into accurate occupancy predictions over the next $T$ frames.

\definecolor{aliceblue}{rgb}{0.94, 0.97, 1.0}
\begin{table*}[ht]
\setlength{\tabcolsep}{0.0140\linewidth}
\newcommand{\graycell}{\cellcolor{gray!10}}
\caption{4D occupancy forecasting performance on Occ3D-nuScenes \cite{occ3d}. Input denotes the modality of input apart from the ego trajectory. Avg. denotes average performance of that in 1s, 2s, and 3s.}
\label{tab:main}
\vspace{-1.0em}
\begin{center}
\scalebox{0.95}{
\begin{tabular}{c|c|cccc|cccc|c}
\toprule
\multirow{2}*{Method} & \multirow{2}*{Input} & \multicolumn{4}{c|}{Semantic mIoU ($\%$) $\uparrow$} & \multicolumn{4}{c|}{Geometric IoU ($\%$) $\uparrow$} & \multirow{2}*{FPS $\uparrow$}\\
\cmidrule(lr){3-6} \cmidrule(lr){7-10}
                   &                             & 1s    & 2s    & 3s    & \graycell Avg.   & 1s    & 2s    & 3s    & \graycell Avg. & \\
\midrule
        
OccWorld-O \cite{occworld}               & Occ GT    & 25.78 & 15.14 & 10.51 & \graycell 17.14     & 34.63 & 25.07 & 20.18 & \graycell 26.63 & 18.00 \\
OccLLaMA-O \cite{occllama}               & Occ GT    & 25.05 & 19.49 & 15.26 & \graycell 19.93     & 34.56 & 28.53 & 24.41 & \graycell 29.17 & - \\
RenderWorld \cite{renderworld}           & Occ GT    & 28.69 & 18.89 & 14.83 & \graycell 20.89     & 37.74 & 28.41 & 24.08 & \graycell 30.08 & -   \\
Occ-LLM \cite{occ-llm}                   & Occ GT    & 24.02 & 21.65 & 17.29 & \graycell 20.99     & 36.65 & 32.14 & 28.77 & \graycell 35.52 & - \\
DFIT-OccWorld \cite{dfit-occworld}       & Occ GT    & 31.68 & 21.29 & 15.19 & \graycell 22.71     & 40.28 & 31.24 & 25.29 & \graycell 32.27 & - \\
DOME \cite{dome}                         & Occ GT    & 35.11 & 25.89 & 20.29 & \graycell 27.10     & 43.99 & 35.36 & 29.74 & \graycell 36.36 & 6.54 \\
UniScene \cite{uniscene}                 & Occ GT    & 35.37 & 29.59 & 25.08 & \graycell 31.76     & 38.34 & 32.70 & 29.09 & \graycell 34.84 & - \\
I$^{2}$-World \cite{i2world}             & Occ GT    & 47.62 & 38.58 & 32.98 & \graycell 39.73     & 54.29 & 49.43 & 45.69 & \graycell 49.80 & 37.04 \\
\midrule
RenderWorld \cite{renderworld}           & Camera    & 2.83  & 2.55  & 2.37  & \graycell 2.58      & 14.61 & 13.61 & 12.98 & \graycell 13.73 & -  \\
OccWorld-D \cite{occworld}               & Camera    & 11.55 & 8.10  & 6.22  & \graycell 8.62      & 18.90 & 16.26 & 14.43 & \graycell 16.53 & - \\
OccLLaMA \cite{occllama}                 & Camera    & 10.34 & 8.66  & 6.98  & \graycell 8.66      & 25.81 & 23.19 & 19.97 & \graycell 22.99 & - \\
PreWorld \cite{preworld}                 & Camera    & 12.27 & 9.24  & 7.15  & \graycell 9.55      & 23.62 & 21.62 & 19.63 & \graycell 21.62 & - \\
Occ-LLM \cite{occ-llm}                   & Camera    & 11.28 & 10.21 & 9.13  & \graycell 10.21     & 27.11 & 24.07 & 20.19 & \graycell 23.79 & - \\
DFIT-OccWorld \cite{dfit-occworld}       & Camera    & 13.38 & 10.16 & 7.96  & \graycell 10.50     & 19.18 & 16.85 & 25.02 & \graycell 17.02 & - \\
OccTENS-F \cite{occtens}                 & Camera    & 17.17 & 10.38 & 7.82  & \graycell 11.79     & 27.60 & 25.14 & 20.33 & \graycell 24.35 & -  \\
SparseWorld \cite{sparseworld}           & Camera    & 14.93 & 13.15 & 11.51 & \graycell 13.20     & 22.96 & 22.10 & 21.05 & \graycell 22.03 & 8.00 \\ 
DOME-STC \cite{dome}                     & Camera    & 17.79 & 14.23 & 11.58 & \graycell 14.53     & 26.39 & 23.20 & 20.42 & \graycell 23.33 & 2.75  \\
I$^{2}$-World-STC \cite{i2world}         & Camera    & 21.67 & 18.78 & 16.47 & \graycell 18.97     & 30.55 & 28.76 & 26.99 & \graycell 28.77 & 4.21  \\
DOME-F \cite{dome}                       & Camera    & 24.12 & 17.41 & 13.24 & \graycell 18.25     & 35.18 & 27.90 & 23.44 & \graycell 28.84 & -  \\
DTT \cite{dtt}                           & Camera    & 24.87 & 18.30 & 15.63 & \graycell 19.60     & 38.98 & 37.45 & 31.89 & \graycell 36.11 & -  \\
COME \cite{come}                       & Camera    & 26.56 & 21.73 & 18.49 & \graycell 22.26     & 48.08 & 43.84 & 40.28 & \graycell 44.07 & -  \\
\rowcolor{aliceblue}
SparseWorld-TC-Small (Ours)                     & Camera  & 27.95 & 25.51 & 23.35 & \graycell 25.60  & 50.69 & 49.15 & 47.23 & \graycell 49.02 & \textbf{9.35}  \\
\rowcolor{aliceblue}
SparseWorld-TC-Large (Ours)                     & Camera  & 28.64 & 26.28 & 24.36 & \graycell 26.42  & 50.57 & 49.26 & 47.80 & \graycell 49.21 & 3.58 \\
\rowcolor{aliceblue}
SparseWorld-TC-Small* (Ours)                    & Camera  & 30.09 & 27.61 & 25.43 & \graycell 27.71  & 52.66 & 51.20 & 49.31 & \graycell 51.05 & 5.63  \\
\rowcolor{aliceblue}
SparseWorld-TC-Large* (Ours)  & Camera & \textbf{32.76} & \textbf{29.62} & \textbf{27.28} & \graycell \textbf{29.89} & \textbf{55.28} & \textbf{53.56} & \textbf{51.71} & \graycell \textbf{53.52} & 2.08 \\
\bottomrule
\end{tabular}
}
\vspace{-0.5em}
\end{center}
\end{table*}

\subsection{Training Strategies}
While the nuScenes occupancy world-model benchmark evaluates forecasting horizons of 1–3 seconds, some studies consider much longer futures (e.g., up to 10 seconds). This motivates us to design a flexible model that adapts to diverse forecasting requirements, supporting future occupancy prediction over arbitrary horizons and even at varying time intervals. We achieve this through a random ensemble strategy that enhances the generalization capability of the trained model without modifying the network architecture.

\bfsection{Random Ensemble Strategy} We assume a maximum forecasting horizon $T$. During training, we randomly choose a target sequence length $L$, where $ L \in \{ 2, \dots, T \}$, and supervise the model with the corresponding $L$ future occupancy frames. Because no fixed temporal stride is prescribed, the scene evolution is governed solely by the trajectory embeddings, which encode temporal and positional context. This flexible supervision scheme enables the model to adapt to diverse forecasting requirements and improves overall performance, as shown in our ablation studies.

\bfsection{Loss Term}
We extract the center of each ground-truth occupancy voxel as target points and optimize the Chamfer Distance loss $\mathcal{L}_{CD}$ (Equation~\ref{eq:cd}) to align the predicted point distribution with the target points. This loss function is widely adopted in point cloud processing~\cite{khurana2023point,yang2024visual} and occupancy modeling~\cite{opus}, as it effectively measures the similarity between predicted point clouds $\mathbb{P}$ and target point clouds $\mathbb{P}_g$. This loss thereby supervises the predicted offsets, guiding the initially randomized points toward the precise underlying geometry.
  
\begin{equation}
    \mathcal{L}_{CD} = \frac{1}{|\mathbb{P}|}\sum\limits_{\mathbf{p} \in \mathbb{P}} D(\mathbf{p}, \mathbb{P}_g) + \frac{1}{|\mathbb{P}_g|}\sum\limits_{\mathbf{p}_g \in \mathbb{P}_g} D(\mathbf{p}_g, \mathbb{P}), 
    \label{eq:cd}
\end{equation}
where $D(\mathbf{x}, \mathbb{Y}) = \min_{\mathbf{y} \in \mathbb{Y}}||\mathbf{x} - \mathbf{y}||_1 $.
Following \cite{opus}, the matched target point also provides a semantic label. Consequently, we supervise semantic predictions using the standard focal classification loss $\mathcal{L}_{focal}$ \cite{focal}, which yields the overall objective:
\begin{equation}
    \label{eq:loss}
    \mathcal{L} = \mathcal{L}_{CD} + \mathcal{L}_{focal}.
\end{equation}

\section{Experiments}
\label{sec:experiments}

\begin{table*}[ht]
\newcommand{\graycell}{\cellcolor{gray!10}}
\caption{Long-term 4D occupancy forecasting performance.}
\label{tab:long-term}
\vspace{-1.0em}
\begin{center}
\small
\renewcommand{\arraystretch}{0.8}
\begin{tabular}{c|c|ccccccccc}
\toprule
\multirow{2}*{Method} & \multirow{2}*{Input} & \multicolumn{9}{c}{Semantic mIoU ($\%$) $\uparrow$} \\
\cmidrule(lr){3-11}
                   &               & 1s    & 2s    & 3s    & 4s    & 5s    & 6s    & 7s    & 8s    & \graycell Avg. \\
\midrule
DOME \cite{dome}   & Occ GT        & 30.10 & 21.35 & 17.36 & 14.86 & 12.61 & 11.03 & 10.00 & 9.34  & \graycell 15.83 \\
COME \cite{come}   & Occ GT        & \textbf{33.78} & 24.57 & 21.35 & 18.25 & 15.84 & 13.85 & 12.99 & 11.96 & \graycell 19.07 \\
SparseWorld-TC-Large  & Camera     & 28.64 & \textbf{26.28} & \textbf{24.36} & \textbf{22.65} & \textbf{21.07} & \textbf{19.73} & \textbf{18.52} & \textbf{17.42} & \graycell \textbf{22.33} \\
\midrule
\multirow{2}*{Method} & \multirow{2}*{Input} & \multicolumn{9}{c}{Geometric IoU ($\%$) $\uparrow$} \\
\cmidrule(lr){3-11}
                   &               & 1s    & 2s    & 3s    & 4s    & 5s    & 6s    & 7s    & 8s    & \graycell Avg. \\
                   \midrule
DOME \cite{dome}   & Occ GT       & 39.04 & 31.20 & 27.14 & 24.73 & 22.32 & 20.28 & 19.05 & 17.97 & \graycell 25.21 \\
COME \cite{come}   & Occ GT       & 44.20 & 36.25 & 32.86 & 30.03 & 26.93 & 24.70 & 23.30 & 21.44 & \graycell 29.96 \\
SparseWorld-TC-Large  & Camera     & \textbf{50.57} & \textbf{49.26} & \textbf{47.80} & \textbf{46.20} & \textbf{44.57} & \textbf{42.98} & \textbf{41.44} & \textbf{39.97} & \graycell \textbf{45.35} \\
\bottomrule
\end{tabular}
\vspace{-1.0em}
\end{center}
\end{table*}

\subsection{Experimental Setup}
\bfsection{Dataset and Metrics} Experiments are conducted on the widely adopted Occ3D-nuScenes benchmark \cite{occ3d}. Our implementation strictly follows the experimental settings established in previous works \cite{dome, occworld, dfit-occworld, come, occ-llm, i2world}. For evaluation, we adopt the standard geometric Intersection over Union (IoU) and semantic mean Intersection over Union (mIoU) metrics. 

\bfsection{Implementation Details} As detailed in the methodology section, our architecture includes several configurable components. We pick past frames within the past 2 seconds, following recent models~\cite{dome, occ-llm, dfit-occworld, occworld, i2world, come}. The future prediction horizon $T$ is set to 3\text{s} and 8\text{s}, aligning with established benchmarks~\cite{occworld,dome}. To accommodate varying computational constraints, we configure the number of anchors per frame as $N = 600$, with each anchor comprising $M = 128$ points. This configuration, designated as SparseWorld-TC-Small, adopts a ResNet-50 backbone \cite{resnet}. Our SparseWorld-TC-Large variant uses $N = 4800$ and $M = 16$, delivering improved performance at increased computational cost. Furthermore, leveraging the representational power of DINOv3~\cite{dinov3}, we introduce SparseWorld-TC-Small* and SparseWorld-TC-Large*, which adopts a DINOv3-Base image backbone while retaining the same configurations as ``-Small'' and ``-Large'' versions. All models are trained for 70 epochs on 8 NVIDIA H20 GPUs with a total batch size of 8, using the AdamW optimizer~\cite{adamw} with an initial learning rate of $2 \times 10^{-4}$ and a cosine annealing learning rate schedule~\cite{lr}.

\subsection{Main Results}
\bfsection{4D Occupancy Forecasting} We evaluate our proposed method following established evaluation protocols~\cite{dome, occworld, dfit-occworld, i2world}, as summarized in Table~\ref{tab:main}. While some VAE-based methods report performance using ground-truth occupancy as input, a requirement sometimes unavailable in real-world scenarios, we include these results for a comprehensive comparison. Since our approach directly consumes camera sensor input rather than precomputed occupancy grids, we primarily compare against camera-based versions of prior methods. Note that OccWorld-D uses predictions from TPVFormer~\cite{tpvformer}; ``-STC'' and ``-F'' indicate models incorporating occupancy results from STCOcc~\cite{stcocc} and FB-Occ~\cite{fbocc}, respectively.

Our SparseWorld-TC-Large achieves significant improvements, outperforming the previous state-of-the-art by 18.7\% in mIoU (26.42 vs. 22.26) and 11.7\% in IoU (49.21 vs. 44.07). Due to the flexible and variable number of queries, our model exhibits strong scalability. SparseWorld-TC-Small achieves 25.60 mIoU and 49.02 IoU, while being approximately twice as fast as SparseWorld-TC-Large and I$^{2}$-World~\cite{i2world}, thus striking an effective balance between performance and efficiency. The strong performance of the DINO version of our method further demonstrates the scalability in leveraging large-scale foundation models.

Moreover, although Table~\ref{tab:main} shows that methods using ground-truth occupancy generally achieve better performance than those processing raw sensor inputs, our camera-based model remains competitive across several metrics. Specifically, SparseWorld-TC-Large* achieves a higher average mIoU (29.89 vs. 27.10) than DOME~\cite{dome} and a higher average IoU (53.52 vs. 49.80) compared to I$^{2}$-World~\cite{i2world}, which can be attributed to its reduced decay in long-term occupancy forecasting.

\begin{figure*}[ht]
  \centering
   \includegraphics[width=1.0\linewidth]{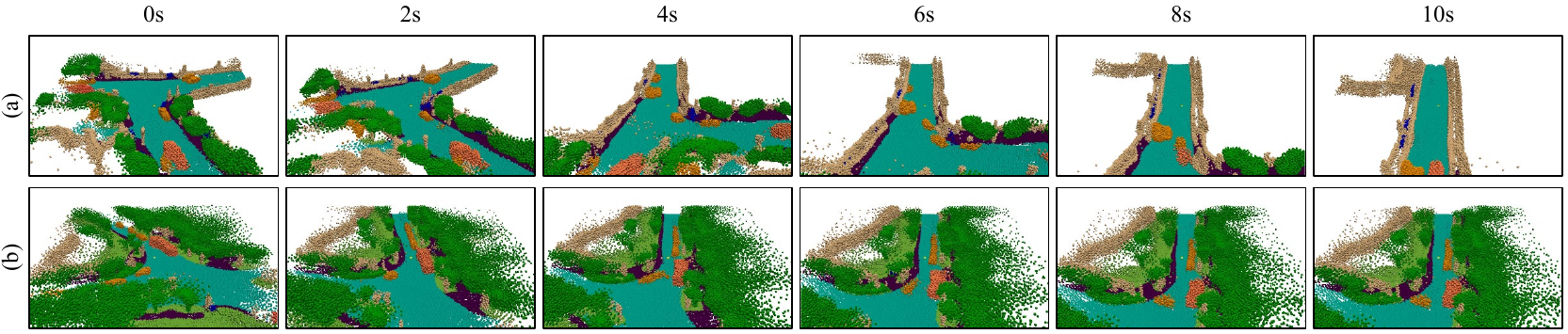}
   \caption{Long-term 4D occupancy world model forecasting.}
   \label{fig:long-term}
   \vspace{-1.0em}
\end{figure*}

\bfsection{Long-term Forecasting}
To evaluate the capability of long-term predictions, we extend the prediction period from 3 seconds to 8 seconds. As shown in Table \ref{tab:long-term}, our method consistently outperforms DOME \cite{dome} and COME \cite{come} after 2 seconds for the mIoU metric while having better performance in all timestamps for the IoU metric. Our method achieves average mIoU and IoU scores of 22.33 and 45.35, surpassing COME \cite{come} by 17.1$\%$ and 51.4$\%$, respectively. It is worth noting that our method uses the camera, while the contrast methods use ground truth occupancy as input.

\subsection{Ablation Study}

\begin{table}[ht]
\newcommand{\graycell}{\cellcolor{gray!10}}
\caption{Ablation study on trajectory. Avg. denotes average performance of mIoU or IoU in 1s, 2s, and 3s.}
\vspace{-1.0em}
\label{tab:ablation}
\begin{center}
\scalebox{0.85}{
\begin{tabular}{ccc|c|c}
\toprule
 \multicolumn{3}{c|}{Trajectory} &  mIoU ($\%$) $\uparrow$ &  IoU ($\%$) $\uparrow$ \\
\cmidrule(lr){1-3} \cmidrule(lr){4-5}
w/o Traj. & Pred. Traj. & GT Traj.                         &  Avg.   &  Avg. \\
\midrule
$\checkmark$  & &   &  15.44           &  32.19  \\
 & $\checkmark$ &   &  21.57           &  44.76  \\
 &  & $\checkmark$  &  \textbf{25.60}  &  \textbf{49.02} \\

\bottomrule
\end{tabular}
}
\vspace{-1.0em}
\end{center}
\end{table}

\bfsection{Effect of Trajectory} Our model supports the conditioning of arbitrary future trajectories. Through ablation studies, we examine how such trajectory guidance influences model behavior and output quality. As summarized in Table~\ref{tab:ablation}, we evaluate performance in three settings based on the SparseWorld-TC-Small model: without trajectory input, with trajectories predicted by BEVPlanner \cite{bev-planner}, and with ground-truth trajectories. The results demonstrate that our model remains robust even when provided with predicted trajectories, while performance consistently improves with more accurate trajectory estimates.


\begin{table}[ht]
\newcommand{\graycell}{\cellcolor{gray!10}}
\caption{Ablation study on training strategies. Avg. denotes average performance of mIoU or IoU in 1-8s.}
\vspace{-1.0em}
\label{tab:rand_ensemble}
\begin{center}
\scalebox{0.85}{
\begin{tabular}{cc|c|c}
\toprule
 \multicolumn{2}{c|}{Training Strategies} &  mIoU ($\%$) $\uparrow$ &  IoU ($\%$) $\uparrow$ \\
\cmidrule(lr){1-2} \cmidrule(lr){3-4}
Fixed & Random Ensemble                                 &  Avg.   &  Avg. \\
\midrule

$\checkmark$ &                                          &  20.36  &  43.25 \\
             & $\checkmark$                             &  \textbf{22.33}  &  \textbf{45.35} \\

\bottomrule
\end{tabular}
}
\vspace{-1.0em}
\end{center}
\end{table}

\bfsection{Effect of Ensemble Training} We explore the impact of our random ensemble training on the SparseWorld-TC-Large model, as shown in Table \ref{tab:rand_ensemble}. In the long-term occupancy forecasting task, the supervision of random future frames is more effective than the supervision of fixed future frames.

\subsection{Visualizations \& Discussions}

\bfsection{4D Occupancy Forecasting}
As shown in Figure~\ref{fig:vis_nusc}, we provide a visual comparison between our reconstruction results, 1-3 second future predictions, and the corresponding ground truth. Qualitatively, our method demonstrates stable and accurate predictions for static environments while also effectively capturing the motion of dynamic objects. In the left-turn scenario depicted in Figure~\ref{fig:vis_nusc}(a), our approach correctly predicts the future occupancy observation of a turning vehicle behind. Similarly, as shown in Figure~\ref{fig:vis_nusc}(b), future occupancy predictions for both straight-moving and left-turning vehicles are reliably predicted.

\begin{figure}[ht]
  \centering
   \includegraphics[width=1.0\linewidth,height=60mm]{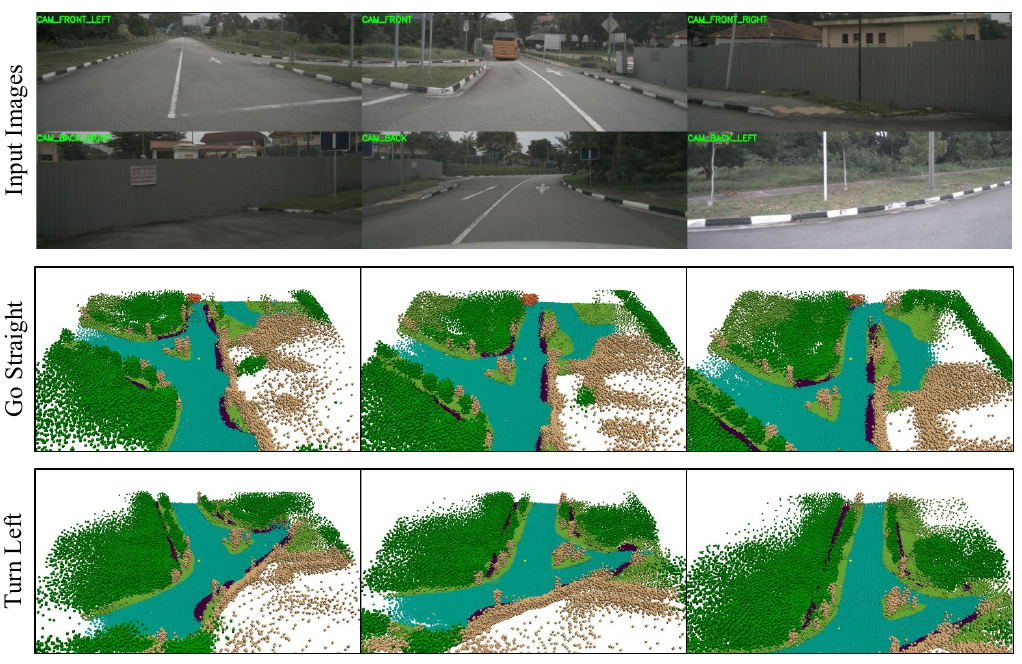}
   \caption{Forecasting with conditioned trajectories.}
   \label{fig:traj_ctrl}
   \vspace{-1.0em}
\end{figure}

\bfsection{Long-term Forecasting}
Figure~\ref{fig:long-term} presents the visualization of our prediction results over a long-term horizon of 10 seconds in two distinct scenarios: (a) a right-turn and (b) a left-turn. Our model maintains strong scene consistency throughout the extended prediction period, with well-preserved geometric structures. The quantitative result is listed in Table~\ref{tab:long-term}. Although the intrinsic multi-possibility of long-term forecasting raises questions about the fairness of using a single ground truth for evaluation, we argue that this metric, combined with the visualization results, at least partially demonstrates the potential of our method for relatively long-term occupancy forecasting.

\bfsection{Trajectory-conditioned Forecasting}
Figure \ref{fig:traj_ctrl} illustrates that the prediction of future driving scenarios is generated by different trajectories as conditions in a bifurcating road scenario, embodying the future occupancy prediction capability conditioned by the trajectory of our model. 

\bfsection{From Occupancy to Sensors} Building upon the occupancy world model, our proposed architecture is further capable of directly predicting future sensor observations, adopting an extension inspired by \cite{infinicube, scube, drivingforward}. This capability is realized by incorporating additional MLPs into the decoding network illustrated in Figure~\ref{fig:representation}. All of these MLPs collectively generate 3D Gaussian parameters, center offsets, and classification logits in a single feed-forward pass. For a comprehensive exposition of our methodology and additional experimental results, please refer to the supplementary document.


\section{Conclusion}
\label{sec:conclusion}

We propose a novel trajectory-conditioned architecture for end-to-end 4D occupancy forecasting, which eliminates the need for discrete tokenization or explicit BEV representations. Using fully sparse anchors and a transformer-based fusion mechanism, our model captures spatiotemporal dependencies directly from sensor data, enabling scalable and robust scene prediction. The flexible design supports the integration of alternative 3D representations such as 3D Gaussians, and exploring these alternatives constitutes a promising direction for future work.


\section*{Acknowledgments}
This paper is supported by the National Natural Science Foundation of China under Grants 62233013. 
{
    \small
    \bibliographystyle{ieeenat_fullname}
    \bibliography{main}
}

\clearpage
\maketitlesupplementary

\renewcommand\thesection{\Alph{section}}
\setcounter{section}{0}
\setcounter{page}{1}

\section{Additional Quantitative Experiments}
\subsection{Ray-level mIoU}
SparseOcc proposes RayIoU (Ray-level mIoU) to solve the inconsistency penalty along the depth axis raised in traditional voxel-level mIoU criteria. We evaluate our SparseWorld-TC-Large* model with this ray-level metric and report the results in Table \ref{tab:riou}.

\begin{table}[ht]
\newcommand{\graycell}{\cellcolor{gray!10}}
\caption{RayIoU scores [$\%$] of 4D occupancy forecasting performance on the Occ3D-nuScenes \cite{occ3d} benchmark.}
\vspace{-1.0em}
\label{tab:riou}
\begin{center}
\scalebox{0.9}{
\begin{tabular}{c|ccc|c}
\toprule
 & RayIoU$_{1m}$ & RayIoU$_{2m}$ & RayIoU$_{4m}$ & RayIoU \\
\midrule
Recon. & 35.4 & 43.0 & 47.8 & 42.1 \\ 
1s     & 29.5 & 36.5 & 41.4 & 35.8 \\
2s     & 25.8 & 32.1 & 36.8 & 31.6 \\
3s     & 23.5 & 29.4 & 33.8 & 28.9 \\

\bottomrule
\end{tabular}
}
\vspace{-1.0em}
\end{center}
\end{table}

\subsection{Ablation Study on Trajectory Embedding}
We further explore the influence of embedding modules for the trajectory condition. The modules named ``TE'', ``PE'' and ``STE'' represent time embedding, position embedding and spatiotemporal embedding, respectively. The performance of our SparseWorld-TC-Small model in different settings is summarized in Table \ref{tab:ablation_modules}.

\begin{table}[ht]
\newcommand{\graycell}{\cellcolor{gray!10}}
\caption{Ablation study on embedding modules. Avg. denotes average performance of mIoU or IoU in 1s, 2s, and 3s.}
\vspace{-1.0em}
\label{tab:ablation_modules}
\begin{center}
\scalebox{0.98}{
\begin{tabular}{ccc|c|c}
\toprule
 \multicolumn{3}{c}{Modules} &  mIoU ($\%$) $\uparrow$ &  IoU ($\%$) $\uparrow$ \\
\cmidrule(lr){1-3} \cmidrule(lr){4-5}
TE & PE & STE                                             &  Avg.   &  Avg. \\
\midrule
   &    &                                                 &  15.44  &  32.19 \\
$\checkmark$ & &                                          &  17.45  &  35.06 \\  
$\checkmark$ & $\checkmark$ &                             &  23.07  &  47.53 \\   
$\checkmark$ & $\checkmark$ & $\checkmark$                &  \textbf{25.60}  &  \textbf{49.02} \\
\bottomrule
\end{tabular}
}
\vspace{-1.0em}
\end{center}
\end{table}

\subsection{Ablations of More Model Settings}
In Table \ref{tab:traj_embed}, ``w/o Deformable Attn'' replaces deformable attention with a simpler multi-view feature aggregation, while ``w/o Temporal Attn'' replaces temporal attention with frame-wise processing. We further ablate the effect of increasing number of anchors and points per anchor in Table \ref{tab:gpu}. Roughly, more anchors and points per anchor both improve the performance. 

\begin{table}[ht]
\newcommand{\graycell}{\cellcolor{gray!10}}
\caption{Ablation on Removing Temporal and Deformable Attention. Avg. denotes average of mIoU or IoU in 1s, 2s, and 3s.}
\vspace{-1.0em}
\label{tab:traj_embed}
\begin{center}
\scalebox{0.9}{
\begin{tabular}{c|c|c}
\toprule
Modules &  Avg.mIoU ($\%$) $\uparrow$ &  Avg.IoU ($\%$) $\uparrow$ \\
\hline
  Ours-Small      &  \textbf{25.60}  &  \textbf{49.02} \\
w/o Temporal Attn.             & 25.07   & 47.53  \\
w/o Deformable Attn.              & 23.89   & 46.55  \\
\bottomrule
\end{tabular}
}
\vspace{-1.0em}
\end{center}
\end{table}

\subsection{Hardware Cost versus Model Size and Performance}
As shown in Table \ref{tab:gpu}, the + denotes increasing the number of anchors (N) or points per anchor (M). We compare on backbones with both ResNet and DINOv3(*).
\begin{table}[ht]
\newcommand{\graycell}{\cellcolor{gray!10}}
\caption{GPU Memory Usage and Performance of Evaluation.}
\vspace{-1.0em}
\label{tab:gpu}
\begin{center}
\scalebox{0.9}{
\begin{tabular}{c|cc|cc}
\toprule
 Model & N & M & Memory (GB) & mIoU ($\%$) \\
\midrule
Ours-Small   & 600 & 128                &  5.4      & 25.60  \\
Ours-Large   & 4800 & 16                &  7.7      & 26.42  \\
Ours-Large+  & 4800 & 64                &  9.2      & 27.25  \\
Ours-Small*  & 600 & 128                &  7.1      & 27.71  \\
Ours-Large*  & 4800 & 16                & 14.4      & 29.89  \\
Ours-Large*+ & 9600 & 16                & 24.6      & 30.88  \\
\bottomrule
\end{tabular}
}
\vspace{-1.0em}
\end{center}
\end{table}

\subsection{Per-class Performance}
As shown in Table \ref{tab:occ_cls_miou} and Table \ref{tab:occ_cls_riou}, we report the performance per-class of our SparseWorld-TC-Large* (Ours-Large* for short). Our approach not only maintains the geometric consistency of static scenes, but also predicts the dynamic objects relatively accurately.

\subsection{Improvement on Small Targets}
The performance on smaller objects, including motorcycles and pedestrians, improves with a larger number of anchor queries. Emphasizing small objects through loss re-weighting also enhances the model capability, as evidenced in Table \ref{tab:small_obj}.

\begin{table}[ht]
\newcommand{\graycell}{\cellcolor{gray!10}}
\caption{Performance Improvement for Small Objects}
\vspace{-1.0em}
\label{tab:small_obj}
\begin{center}
\scalebox{0.9}{
\begin{tabular}{c|cc}
\toprule
 \multirow{2}{*}{Model} &  \multicolumn{2}{c}{Avg. mIoU ($\%$) $\uparrow$} \\
 \cmidrule(lr){2-3}
                            & motorcycle & pedestrian \\
\midrule
Ours-Large*                 & 13.64      &  6.86       \\
Q=4800 $\rightarrow$ Q=9600 & 14.72 (+1.08)     &  8.29 (+1.43)      \\  
Class Reweighting           & 14.64 (+1.00)     &  8.63 (+1.83)     \\  
\bottomrule
\end{tabular}
}
\end{center}
\vspace{-1.0em}
\end{table}

\section{Additional Qualitative Experiments}

\subsection{Feedforward Gaussian}
Inspired by advances in feedforward Gaussian methods \cite{scube, infinicube, drivingforward}, we extend the original model with additional MLPs to decode Gaussian parameters from latent features. Then we utilize the differential Gaussian rasterization proposed in 3DGS~\cite{3dgaussian} to render the predicted front-view image and calculate L1 loss with the GT image.

Currently our feedforward 3DGS implementation only supports the front-view camera, which supports both 256x704 and 128x352 resolutions. To alleviate the voids caused by query sparsity, we boost the number of anchors and points per anchor to 7200 and 32.
\begin{figure}[ht]
\centering
\footnotesize
\setlength{\tabcolsep}{0.05cm}
\newcolumntype{P}[1]{>{\centering\arraybackslash}m{#1}}
\scalebox{0.7}{
\begin{tabular}{P{0.7cm}P{5cm}P{5cm}}
 & GT & Pred. \\

\rotatebox[origin=c]{90}{(a)} & {\includegraphics[width=\linewidth, frame]{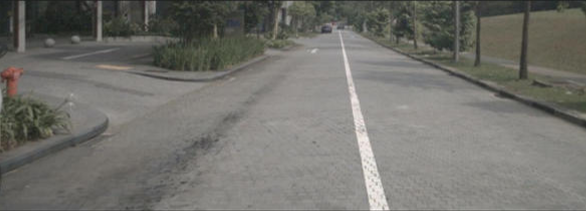}} & {\includegraphics[width=\linewidth, frame]{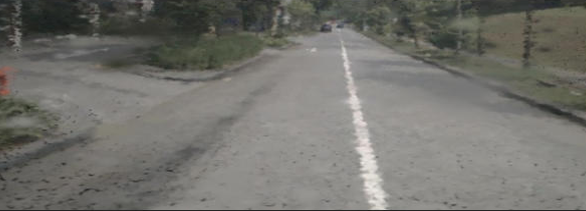}} \\

\rotatebox[origin=c]{90}{(b)} & {\includegraphics[width=\linewidth, frame]{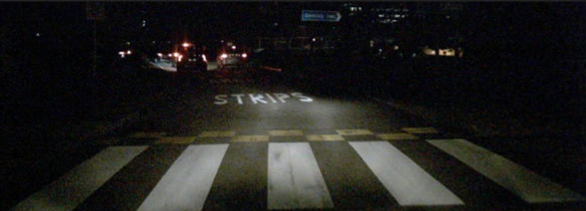}} & {\includegraphics[width=\linewidth, frame]{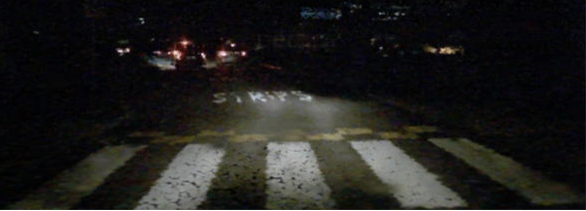}} \\

\rotatebox[origin=c]{90}{(c)} & {\includegraphics[width=\linewidth, frame]{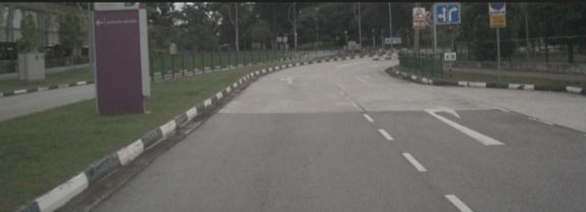}} & {\includegraphics[width=\linewidth, frame]{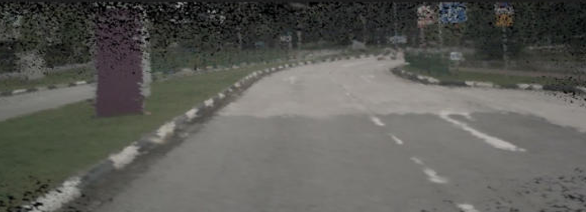}} \\

\rotatebox[origin=c]{90}{(d)} & {\includegraphics[width=\linewidth, frame]{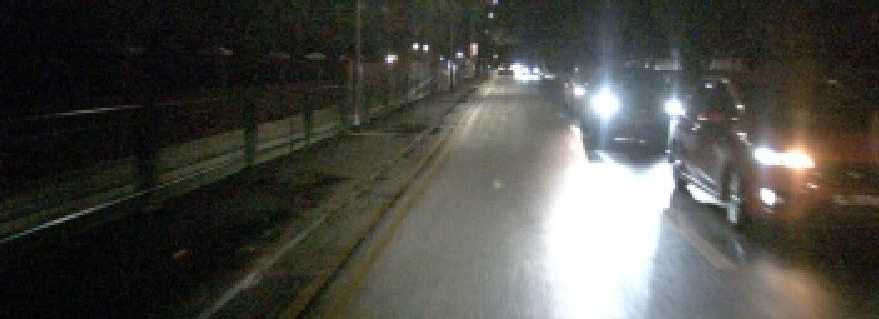}} & {\includegraphics[width=\linewidth, frame]{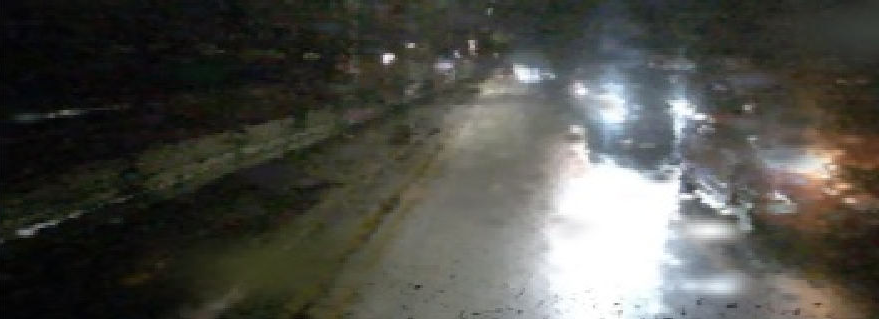}} \\

\rotatebox[origin=c]{90}{(e)} & {\includegraphics[width=\linewidth, frame]{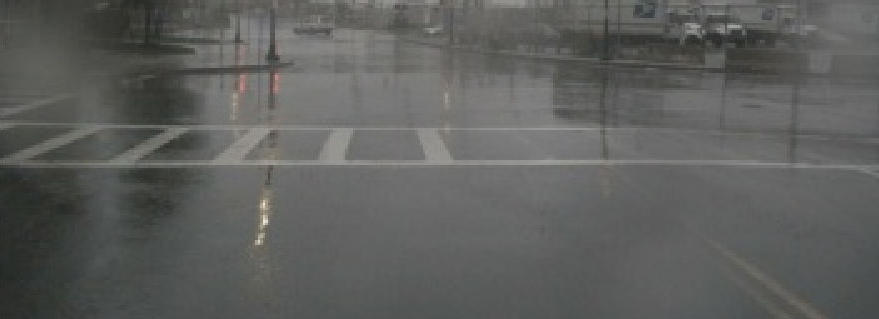}} & {\includegraphics[width=\linewidth, frame]{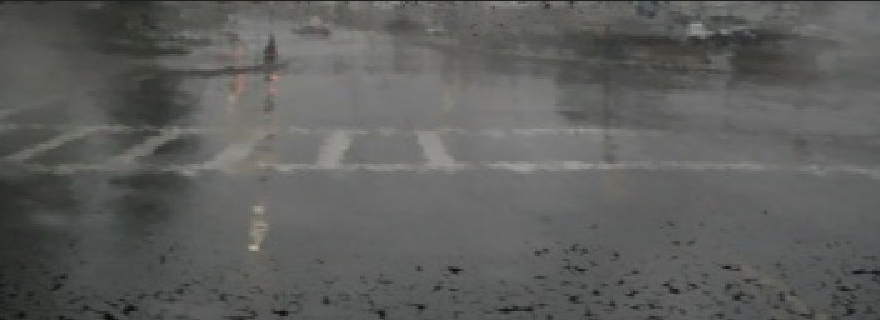}} \\

\end{tabular}
}
\caption{Gaussian splatting reconstruction during training.}
\vspace{-1.0em}
\label{fig:vis_gs_train}
\end{figure}

\begin{figure}[ht]
\centering
\footnotesize
\setlength{\tabcolsep}{0.05cm}
\newcolumntype{P}[1]{>{\centering\arraybackslash}m{#1}}
\scalebox{0.7}{
\begin{tabular}{P{0.7cm}P{5cm}P{5cm}}
 & GT & Pred. \\

\rotatebox[origin=c]{90}{1s} & {\includegraphics[width=\linewidth, frame]{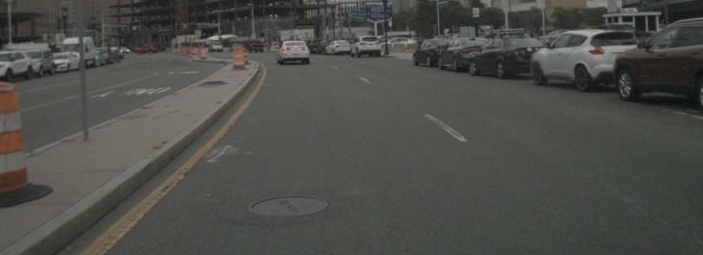}} & {\includegraphics[width=\linewidth, frame]{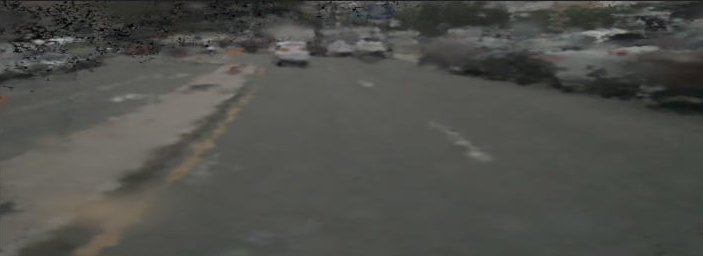}} \\

\rotatebox[origin=c]{90}{2s} & {\includegraphics[width=\linewidth, frame]{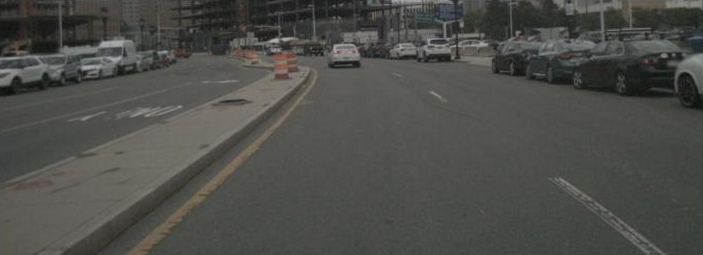}} & {\includegraphics[width=\linewidth, frame]{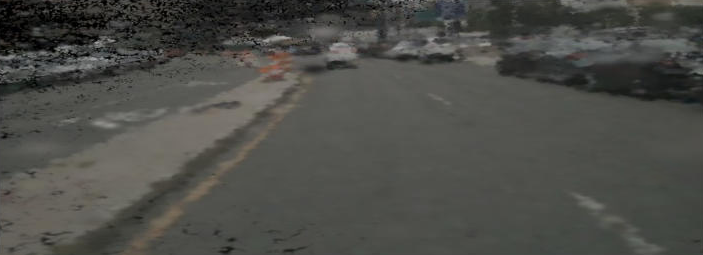}} \\

\rotatebox[origin=c]{90}{3s} & {\includegraphics[width=\linewidth, frame]{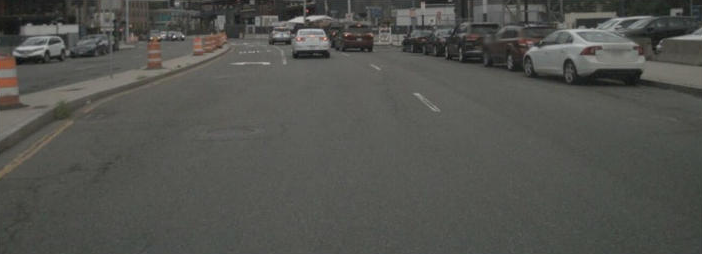}} & {\includegraphics[width=\linewidth, frame]{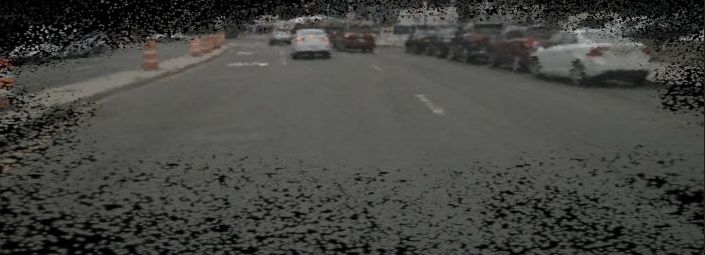}} \\

\end{tabular}
}
\caption{Future observation forecasting on validation set.}
\vspace{-1.0em}
\label{fig:vis_gs_val}
\end{figure}

The reconstruction and future forecasting of sensor observation, as shown Figure \ref{fig:vis_gs_train} and Figure \ref{fig:vis_gs_val},  demonstrate the potential of our model in leveraging the Gaussian representation and achieving self-supervised training in the future.

\subsection{Convergence Visualization}
\begin{figure}[h]
  \centering
  \includegraphics[width=1.0\linewidth]{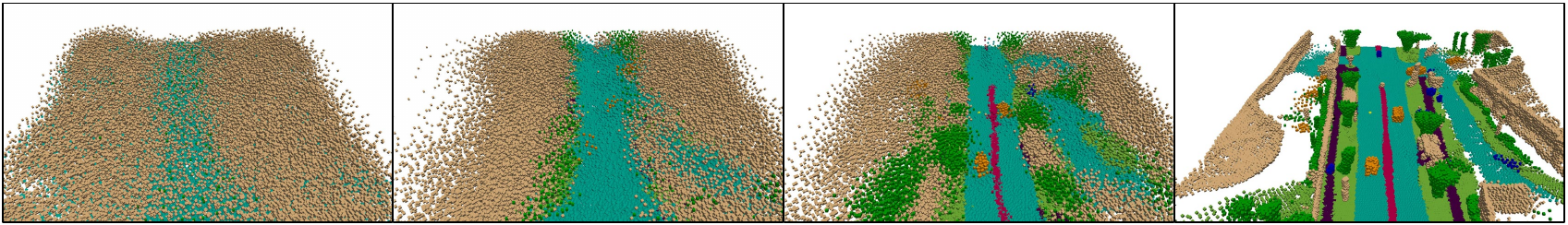}
   \caption{Model Convergence Visualization.}
   \label{fig:iter}
\end{figure}

Figure \ref{fig:iter} shows how the model converge from noise to the detailed occupancy.

\subsection{Failure Case Analysis}
\begin{figure}[h]
  \centering
  \includegraphics[width=0.95\linewidth,height=40mm]{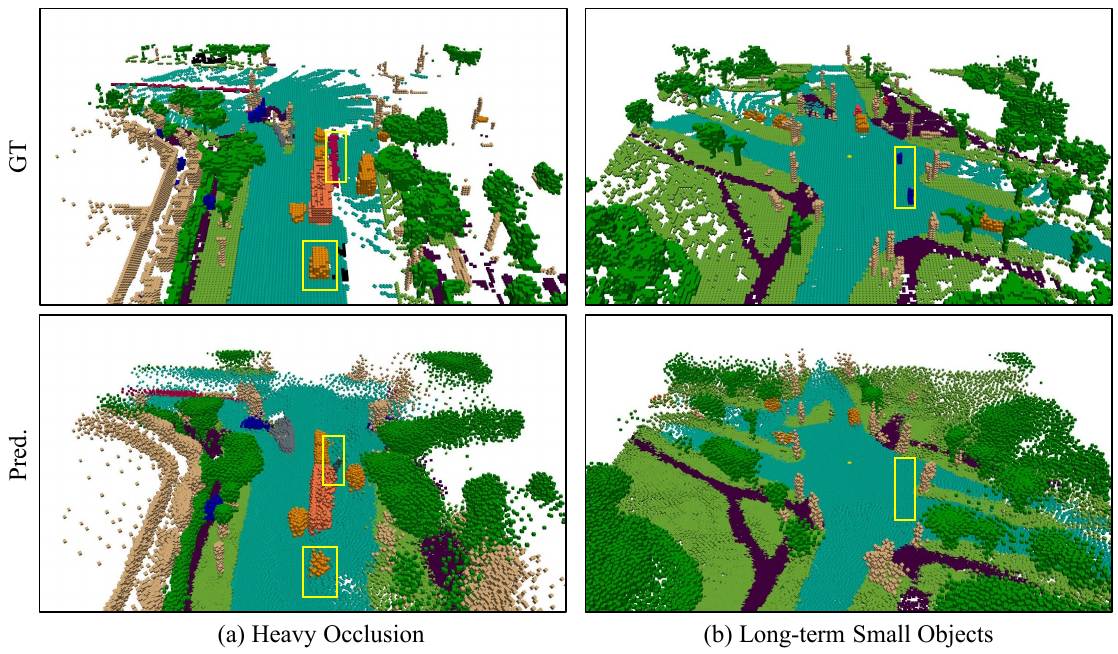}
   \caption{Failure cases: (left) the car and barrier with heavy occlusion, (right) small objects appear in the future 6 seconds.}
   \label{fig:failure}
\end{figure}
As shown in Fig.~\ref{fig:failure}, we visualize typical failure cases.

\subsection{Additional Trajectory-conditioned Prediction}
\begin{figure*}[hb]
  \centering
   \includegraphics[width=1.0\linewidth]{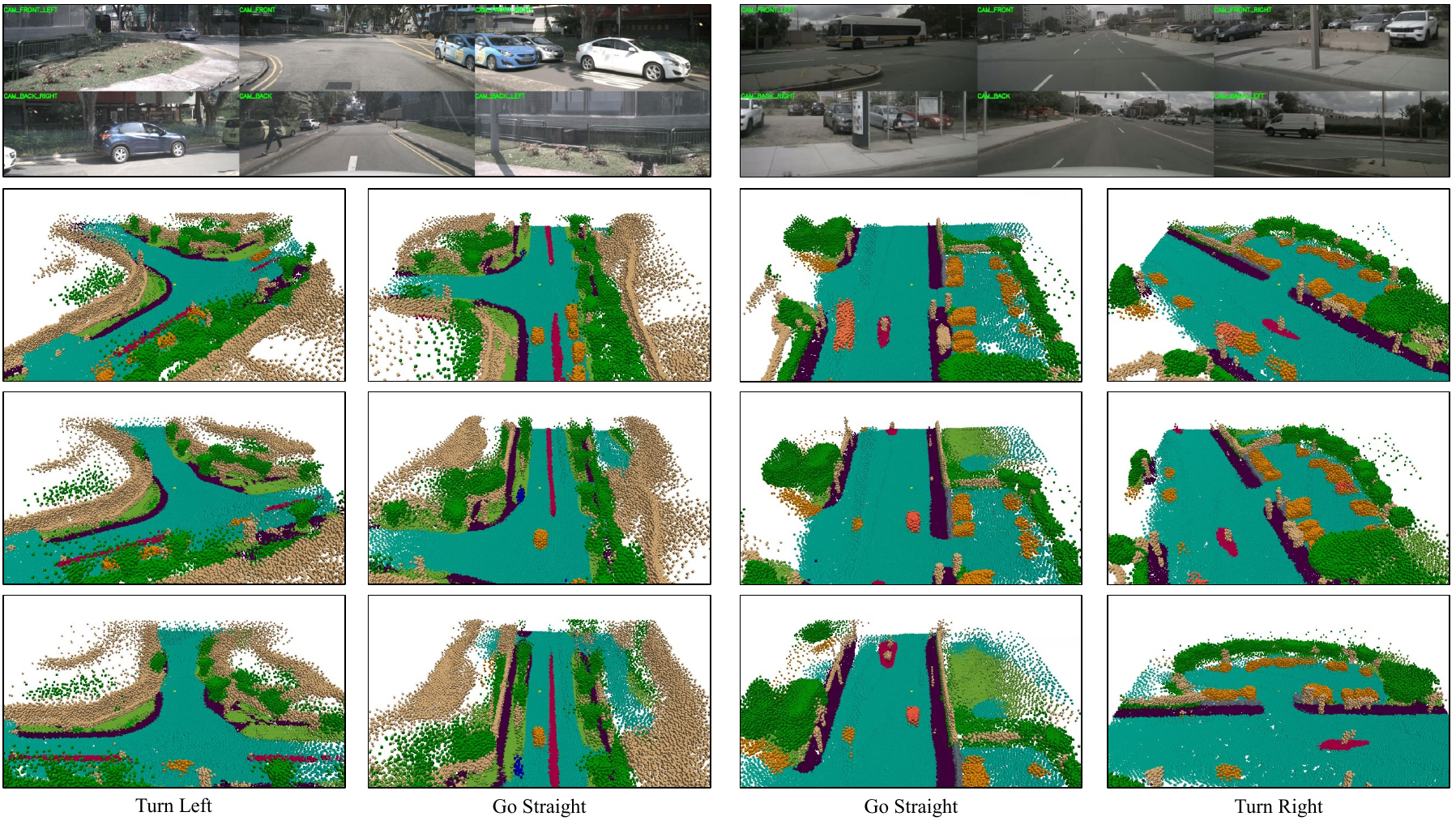}
   \caption{Additional qualitative results in the different trajectory conditions.}
   \label{fig:tc}
\end{figure*}

\begin{figure*}
    \centering
    \includegraphics[width=1.0\linewidth]{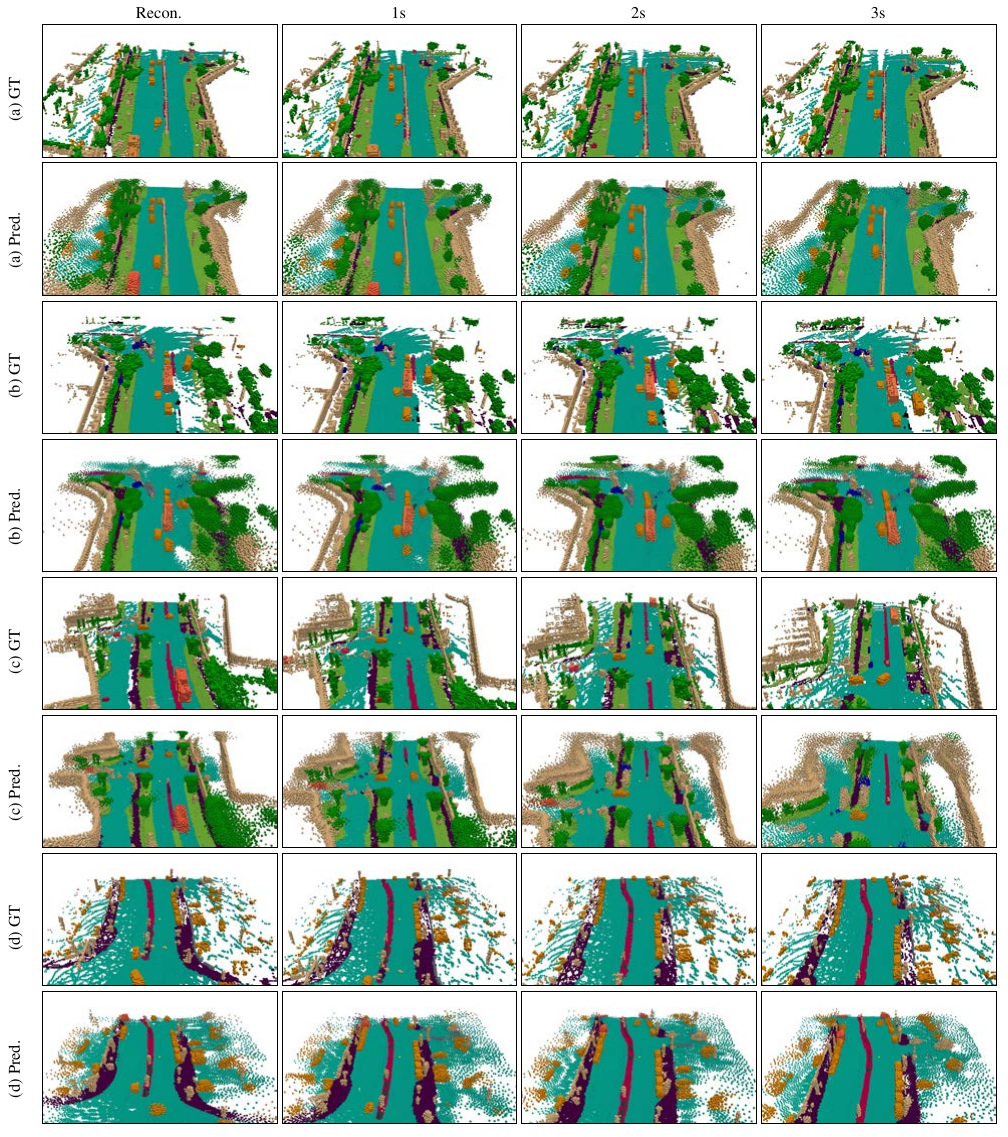}
    \caption{Additional qualitative results of our proposed SparseWorld-TC are presented here.}
    \label{fig:add_vis0}
\end{figure*}

\begin{figure*}[ht]
\centering
\includegraphics[width=1.0\linewidth]{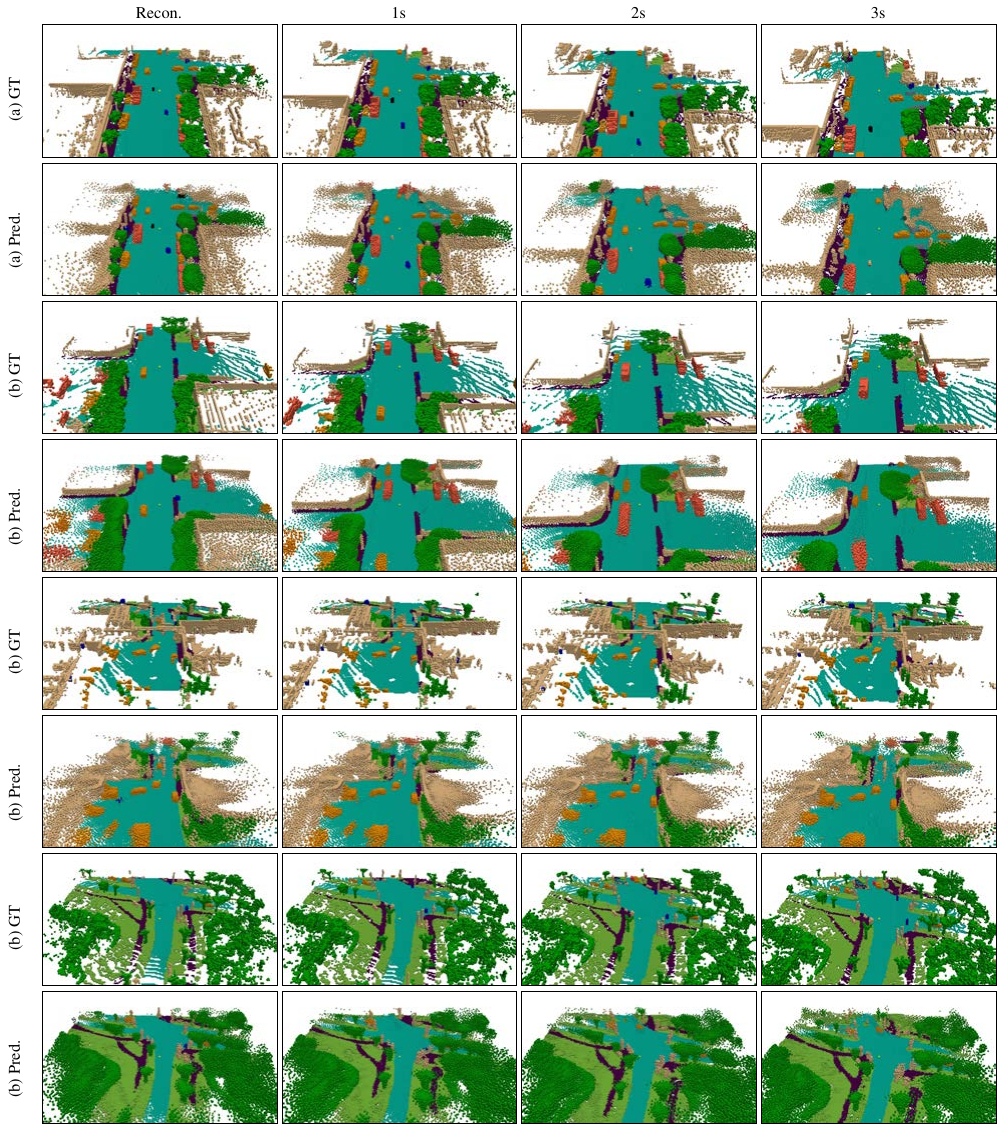}
\caption{Additional qualitative results of our proposed SparseWorld-TC are presented here.}
\label{fig:add_vis1}
\end{figure*}

We additionally visualize the occupancy prediction results under different trajectory conditions, as shown in Fig.~\ref{fig:tc}.
More visualizations are shown in Fig.~\ref{fig:add_vis0} and Fig.~\ref{fig:add_vis1}.

\definecolor{nbarrier}{RGB}{112, 128, 144}
\definecolor{nbicycle}{RGB}{220, 20, 60}
\definecolor{nbus}{RGB}{255, 127, 80}
\definecolor{ncar}{RGB}{255, 158, 0}
\definecolor{nconstruct}{RGB}{233, 150, 70}
\definecolor{nmotor}{RGB}{255, 61, 99}
\definecolor{npedestrian}{RGB}{0, 0, 230}
\definecolor{ntraffic}{RGB}{47, 79, 79}
\definecolor{ntrailer}{RGB}{255, 140, 0}
\definecolor{ntruck}{RGB}{255, 99, 71}
\definecolor{ndriveable}{RGB}{0, 207, 191}
\definecolor{nother}{RGB}{175, 0, 75}
\definecolor{nsidewalk}{RGB}{75, 0, 75}
\definecolor{nterrain}{RGB}{112, 180, 60}
\definecolor{nmanmade}{RGB}{222, 184, 135}
\definecolor{nvegetation}{RGB}{0, 75, 0}
\definecolor{nothers}{RGB}{0, 0, 0}

\begin{table*}[hbt]
    \caption{Per-class mIoU [$\%$] performance of 4D occupancy forecasting on Occ3D-nuScenes \cite{occ3d}.}
    \label{tab:occ_cls_miou}
	\footnotesize
 	\setlength{\tabcolsep}{0.0040\linewidth}
	\newcommand{\classfreq}[1]{{~\tiny(\nuscenesfreq{#1}\%)}}  %
    \begin{center}
	\begin{tabular}{c|c|c| c c c c c c c c c c c c c c c c c}
		\toprule
		\rotatebox{90}{Ours-Large*}
		& \rotatebox{90}{mIoU} & \rotatebox{90}{IoU}
        & \rotatebox{90}{\textcolor{nothers}{$\blacksquare$} others}
		& \rotatebox{90}{\textcolor{nbarrier}{$\blacksquare$} barrier}
		& \rotatebox{90}{\textcolor{nbicycle}{$\blacksquare$} bicycle}
		& \rotatebox{90}{\textcolor{nbus}{$\blacksquare$} bus}
		& \rotatebox{90}{\textcolor{ncar}{$\blacksquare$} car}
		& \rotatebox{90}{\textcolor{nconstruct}{$\blacksquare$} const. veh.}
		& \rotatebox{90}{\textcolor{nmotor}{$\blacksquare$} motorcycle}
		& \rotatebox{90}{\textcolor{npedestrian}{$\blacksquare$} pedestrian}
		& \rotatebox{90}{\textcolor{ntraffic}{$\blacksquare$} traffic cone}
		& \rotatebox{90}{\textcolor{ntrailer}{$\blacksquare$} trailer}
		& \rotatebox{90}{\textcolor{ntruck}{$\blacksquare$} truck}
		& \rotatebox{90}{\textcolor{ndriveable}{$\blacksquare$} drive. suf.}
		& \rotatebox{90}{\textcolor{nother}{$\blacksquare$} other flat}
		& \rotatebox{90}{\textcolor{nsidewalk}{$\blacksquare$} sidewalk}
		& \rotatebox{90}{\textcolor{nterrain}{$\blacksquare$} terrain}
		& \rotatebox{90}{\textcolor{nmanmade}{$\blacksquare$} manmade}
		& \rotatebox{90}{\textcolor{nvegetation}{$\blacksquare$} vegetation}
		\\
		\midrule
Recon. & 37.92 & 57.65 & 13.15 & 45.06 & 26.72 & 40.26 & 46.44 & 24.62 & 25.38 & 21.47 & 28.99 & 34.84 & 35.92 & 76.87 & 44.76 & 51.14 & 49.71 & 40.98 & 38.38      \\
1s     & 32.76 & 55.28 & 12.25 & 42.04 & 19.22 & 29.07 & 32.03 & 22.42 & 15.70 & 10.28  & 24.24 & 30.55 & 27.37 & 74.98 & 43.12 & 49.53 & 48.43 & 39.03 & 36.68      \\
2s     & 29.62 & 53.56 & 11.40 & 38.75 & 15.61 & 20.72 & 25.49 & 19.93 & 13.46 & 6.01  & 19.55 & 26.27 & 22.35 & 74.46 & 42.20 & 48.34 & 47.42 & 36.82 & 34.76      \\
3s     & 27.28 & 51.71 & 10.36  & 35.19 & 12.38  & 16.90 & 22.22 & 17.73 & 11.75  & 4.29  & 15.14 & 23.78 & 20.04 & 73.56 & 40.62 & 46.71 & 45.89 & 34.49 & 32.66      \\
		\bottomrule
	\end{tabular}
    \end{center}
\end{table*}

\begin{table*}[h]
    \caption{Per-class RayIoU [$\%$] performance of 4D occupancy forecasting on Occ3D-nuScenes \cite{occ3d}.}
    \label{tab:occ_cls_riou}
	\footnotesize
 	\setlength{\tabcolsep}{0.0040\linewidth}
	\newcommand{\classfreq}[1]{{~\tiny(\nuscenesfreq{#1}\%)}}  %
    \begin{center}
	\begin{tabular}{c|c| c c c c c c c c c c c c c c c c c}
		\toprule
		\rotatebox{90}{Ours-Large*}
		& \rotatebox{90}{RayIoU}
        & \rotatebox{90}{\textcolor{nothers}{$\blacksquare$} others}
		& \rotatebox{90}{\textcolor{nbarrier}{$\blacksquare$} barrier}
		& \rotatebox{90}{\textcolor{nbicycle}{$\blacksquare$} bicycle}
		& \rotatebox{90}{\textcolor{nbus}{$\blacksquare$} bus}
		& \rotatebox{90}{\textcolor{ncar}{$\blacksquare$} car}
		& \rotatebox{90}{\textcolor{nconstruct}{$\blacksquare$} const. veh.}
		& \rotatebox{90}{\textcolor{nmotor}{$\blacksquare$} motorcycle}
		& \rotatebox{90}{\textcolor{npedestrian}{$\blacksquare$} pedestrian}
		& \rotatebox{90}{\textcolor{ntraffic}{$\blacksquare$} traffic cone}
		& \rotatebox{90}{\textcolor{ntrailer}{$\blacksquare$} trailer}
		& \rotatebox{90}{\textcolor{ntruck}{$\blacksquare$} truck}
		& \rotatebox{90}{\textcolor{ndriveable}{$\blacksquare$} drive. suf.}
		& \rotatebox{90}{\textcolor{nother}{$\blacksquare$} other flat}
		& \rotatebox{90}{\textcolor{nsidewalk}{$\blacksquare$} sidewalk}
		& \rotatebox{90}{\textcolor{nterrain}{$\blacksquare$} terrain}
		& \rotatebox{90}{\textcolor{nmanmade}{$\blacksquare$} manmade}
		& \rotatebox{90}{\textcolor{nvegetation}{$\blacksquare$} vegetation}
		\\
		\midrule
Recon. & 42.1  & 11.7 & 47.3 & 31.7 & 63.6 & 56.3 & 28.7 & 30.4 & 35.2 & 32.3 & 37.1 & 52.4 & 70.4 & 40.7 & 38.1 & 40.2 & 53.1 & 46.2      \\
1s     & 35.8  & 10.5 & 44.8 & 19.8 & 47.2 & 43.0 & 26.7 & 17.6 & 21.6 & 29.1 & 31.0 & 44.5 & 66.0 & 37.9 & 35.5 & 38.3 & 50.8 & 44.2      \\
2s     & 31.6  & 9.7  & 41.9 & 15.9 & 34.4 & 34.1 & 25.1 & 15.1 & 13.7 & 25.8 & 21.9 & 39.7 & 62.8 & 36.1 & 33.6 & 36.7 & 48.3 & 41.8      \\
3s     & 28.9  & 9.0  & 39.3 & 12.9 & 28.5 & 29.9 & 23.9 & 13.5 & 10.1 & 22.5 & 19.1 & 36.7 & 60.2 & 34.1 & 31.8 & 34.8 & 45.8 & 39.3      \\
		\bottomrule
	\end{tabular}
    \end{center}
\end{table*}

\end{document}